# *Exploring high-level Perspectives on Self-Configuration Capabilities of Systems*

*Aleksander Lodwich*

*aleksander[at]lodwich.net*

**Abstract** – Optimization of product performance repetitively introduces the need to make products adaptive in a more general sense. This more general idea is often captured under the term "self-configuration". Despite the importance of such capability, research work on this feature appears isolated by technical domains. It is not easy to tell quickly whether the approaches chosen in different technological domains introduce new ideas or whether the differences just reflect domain idiosyncrasies. For the sake of easy identification of key differences between systems with self-configuring capabilities, I will explore higher level concepts for understanding self-configuration, such as the Ω units, in order to provide theoretical instruments for connecting different areas of technology and research.

## 1 Self-organization, Structural Stabiliy and Self-Configuration

In order to understand how organisms and intelligent systems achieve their remarkable adaptivity, it could be interesting to look at their capabilities from the point of view of abstract systems but it seems to be the case that systemic theories of self-configuration are missing.

A similar area of research is that of self-organization. Self-organization as a phenomenon of structure emergence in dynamical systems has been studied for a long time [1].

Systems with capability to organize themselves are said to be adaptive and robust [2]. Self-organization is very often introduced in terms of phenomena observed in physical systems which are showing remarkable structures[1] under certain conditions. Gershenson [2] explains this phenomenon with a two layer process of entropy transport: Entropy increased in lower levels of a system abstraction can result in decreased entropy in higher level system representations.

Alternatively, this process is explained in terms of a system with feedback to strive for an energy basin. Perturbations of the system are corrected by forces striving back to the basin's center. Yet another explanation is that of dynamic equilibria as were first proposed by Andronov and Pontryagin in [3].

---

1 http://www.scholarpedia.org/article/Self-organization

The study of formation of structures in dynamical systems is sometimes referred to as study of *structural stability*. So to say, study of self-organization studies the process of creating the structures while the study of structural stability concentrates on their resilience. This is a notable sharing of chores suggesting that the original forces of structure formation could become irrelevant at some point of time in a system's evolution and a different set of forces could dominate the structural configuration henceforth. This could lead to autonomy of the structure.

Despite that Gershenson [2] mentions anticipative self-organization, he has only one source to refer to [4] which is a signal simply by the quantity (if compared to references related to adaptive and robustness properties) that anticipative and active organization is not strongly associated to the study of self-organization. One of possible explanations could be that externally enforced configurations do not share properties with organizations obtained from self-organizing processes. A configuring process could consume energy and could yield results which show no clear relationship to dynamic equilibria or transport of entropy. In that case generic theoretical frameworks known from study of self-organization and study of structural stability are of limited utility to the study of self-configuring systems. If we briefly assume that self-organization is more related to self-optimization then we see several sources which clearly see self-configuration and self-optimization (and hence self-organization) as distinct processes [5][6][7][8]. Most explicitly, Reza Nami and Sahrifi associate self-configuration with the idea of portability of functionality – an idea that shows the potential of treating self-configuration beyond topics of structure and organization.

A scan of survey titles on self-configuring systems brought up the conclusion that literature on self-configuring systems seems to be suffering fragmentation along the lines of technical domains and disciplines.

Results were obtained for self-configuration for (computer / signal) networks [5][9], self-configuration for computing substrates [10][11], self-configuration for robotics [12][13] and self-configuration for various pure software and hybrid technologies, such as IoT [14] or databases for big data environments [15]. I exclude references to software configuration management literature because these are concerned with techniques related to building products (i.e. they are not *self*-configurable). But, of course, a systemic understanding of configuration should also provide some understanding for configuration management for systems (product lifecycle management, PLM) and software products (application lifecycle management, ALM).

So far, the cited literature seems to be skipping some kind of middle ground between high level considerations and domain-specific technical implementations: After briefly introducing the expected promises of self-configuration, work quickly skips to questions of mathematical and technical realization. Maybe because it seems trivial in their context but I see some examples where some of middle ground is explored: These are works from Kokar, Badawski and Eracar [16] and from Williams and Pandurang Nayak [17].

In [16], Kokar, Badawski and Eracar propose to understand the self-configuration process and self-configuration capabilities in a broader context of applications. They object that configuration (or self-configuration) is not a mere hardware science and may very well be applied to software domains. Since software applications and software environments are barely ever without temporal evolution (dynamics) they propose to rely on concepts from the control theory in order to explain self-configuring capabilities also in software.

Figure 1 shows a (slightly interpreted) result of Kokar, Badawski and Eracar: In this model the plant is defining the bulk of the behavior of the system but the outer regulator loop to adapt its function is clearly detectable.

The authors attempted to combine fast "small scale" adaptivity with larger "mode changes" which they associate with the term

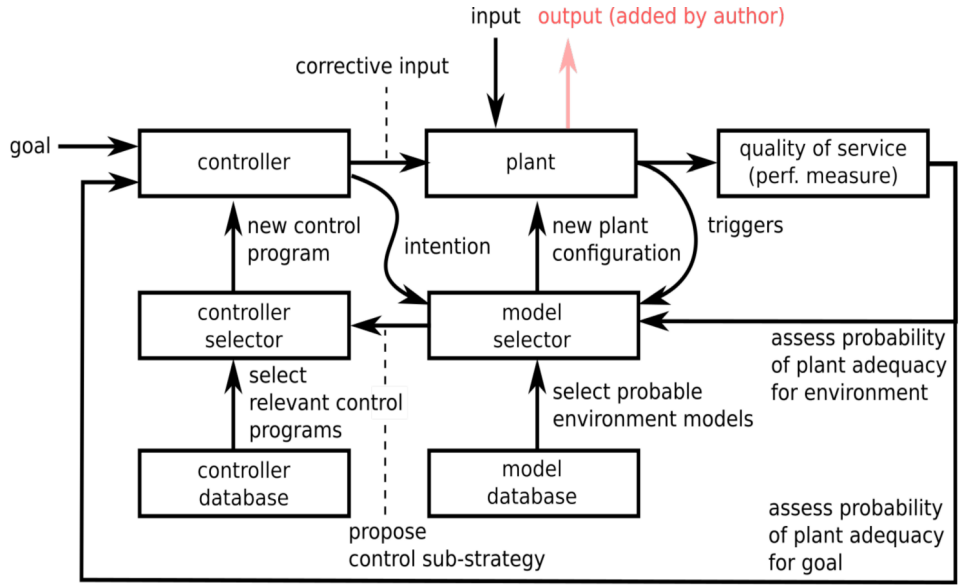

*Figure 1: Kokar, Badawski and Eracar [16] proposed a model for self-configuring systems by expanding typical control theoretical patterns. This schema has been adapted for the purpose of this paper: pink (added by author), letter identifiers of the connections were replaced with intended activities.*

"reconfiguration" or "self-reconfiguration". The reconfigurable *controller* is performing small-scale adaptivity while the major mode changes are realized by two independently reconfiguring components, the *controller selector* and the *model selector*.

In [17], Williams and Pandurang Nayak propose a more compact version of a (re-)configuring control loop for their Livingstone kernel used to control NASA's space mission probes as is shown in figure 2. In this model the *configuration manager* implicitly also takes on the role of the *controller database*, *controller selector* and the *controller* seen in figure 1, implicating that small-scale adaptivity and large-scale adaptivity (associated with multi-modal behavior changing capabilities) can be indeed generalized into a single concept.

There is another difference: The *planner* in figure 2 seems to have no representation in figure 1. The planner is responsible for generating higher level goals based on confirmations received from the configuration manger (dynamic goals) which are simply assumed in figure 1 to be a given input (static goals).

In practical terms, the two architectural styles can be combined in a nesting fashion: For example, the outer system could be made adaptively reconfiguring by the concept of Williams and Pandurang Nayak and the system's plant could include additional adaptivity according to the model of Kokar,

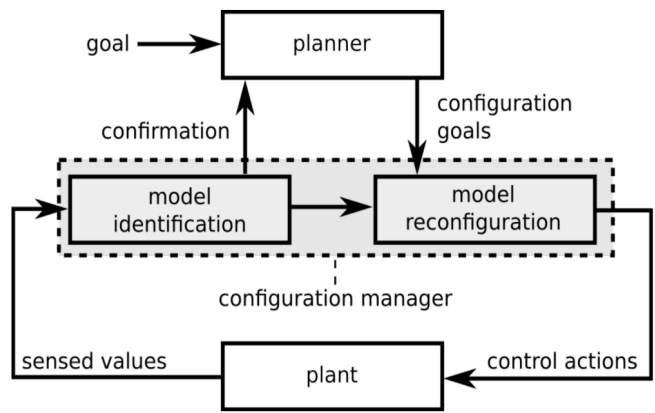

*Figure 2: Model for reconfiguring the Livingstone kernel as described by Williams and Pandurang Nayak in [17].*

Badawski and Eracar. Adaptivity and reconfiguring capabilities of the hypothetical system could be also created by the inverted nesting. It appears also possible to re-nest the same concepts. In that case the question arises if the self-configuring capabilities are optimally arranged and if they can be rationalized. In some cases it could be impossible to rationalize a nested architecture for reconfiguration, for example because such nesting reflects an external system of authorization (security architecture). Authorities over high-level configurations could lack privileges to control lower level functions and *vice versa*.

Since such nestings demand that outputs of one system are input to another, the question arises how configurations, parameters or goals could relate together. In further discourse I will try to boil this terminology down to just a single concept of configuration. With a single concept cascading of self-configuring systems is easier to understand.

Since the two models from literature propose self-configuring systems with modal reconfigurations and with continuous reconfigurations, it is an interesting question to discuss how to choose between the two approaches and if a choice is necessary?

In order to address these question I want to translate previously presented models into an information theoretic concept of a self-configuring system ($\Omega$-unit). Such model shall be independent of applications or chosen technology (systemic) and yet allow conclusions for technical architectures. For this, I will explore the proposition's properties and influence on behavior. Finally, I will draw conclusions for systems designers.

## 2 *Configuration Spaces*

### 2.1 *Introduction to Spaces*

Since current literature does not propose a universal concept of configuration processes for the engineer, and since self-configuration is considered as a feature distinct from other self-capabilities, I would like to propose a systemic, information-based "middle grounds" concept of configuration processes taking place in systems. This concept consists of a components-triple (storage, configurator, plant) which I will call $\Omega$-units for the sake of easy reference.

The model relies a little bit on the idea that configurations have a uniform unit of quantity describing their size – the bit. It is easier to imagine this if different types of configurations could be converted into a single, easily streamable concept – a bit pattern for a point in a configuration space.

In the following sub-chapters I will discuss the term configuration, involved conceptual frameworks and their convertibility in hope that it is possible to use the *$\Omega$-units with configuration spaces* as a general model in which any self-configuring system could be on one hand classified and on second hand assessed in terms of which optimizations could be applied to them.

### 2.2 *On Terminology*

What means configuration or self-configuration? There are several ways to think about those terms, either as a state, capability or a process.

**Configuration as State**: In this sense the meaning is equivalent with "pattern" or "defined pattern", "useful pattern", "collection with defined relationships". This type of semantic is used for example in *configuration management* where the task is to trace down working constellations out of many similar ones which do not work.

**Configuration as Capability**: In this sense we could also speak of "configurability". Lego blocks have such capability. They have defined options of connecting elements in order to form more complex structures. Clearly, Lego blocks can become parts in configurations and resist some forces of disorder. In some overarching perspective this could be seen as *some* kind of self-configuration capability. Configurability does not necessarily imply an active mechanism.

However, brothers in heart to configurability – basically as a passive property of being able "to be put reasonably together" - are interoperability, composability or integratability which are used in scenarios where "being put together" requires some kind of active support on side of the parts [18]. Since this can be considered a prerequisite for any sensible self-configuration, it is not surprising that configurability is important area of research (example from various areas [19][20][20][21][22]).

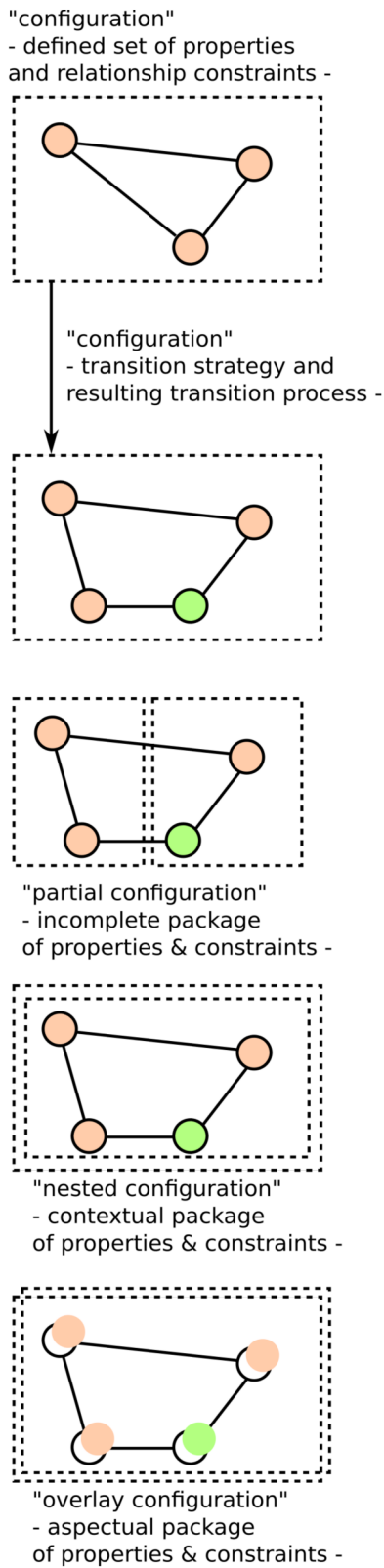

Figure 3: Various ways to speak of configurations.

C*onfiguration as capability* can also appear under such terms as *structure research* or *compatibility*.

**Configuration as Process**: With configuration as process we focus more on the act of going from an unordered state (or not usefully ordered state) into another ordered state (and particularly useful one). The self-configuration capability relies on the ability to efficiently transit from one configuration to another and to avoid access to undesired (for example dangerous or irreversible) configurations.

**Configuration as System of Systems**: A frequent notion in literature is that there exist partial configurations, configurations which need additional pieces of information in order to establish a working configuration, either by a) needing loosely related sibling configuration packages, b) by needing an external "framing" configuration package or c) by requiring "overlaying" configurations.

The visual representations can be found in figure 3. It shows a graph representation of a configuration. The colors indicate certain node parameters. Such a configuration can be at least split, nested or overlaid.

The ability of configurations to be altered, split, nested, overlaid or otherwise factored out shall find direct representations in operations possible with Ω-units. For example, if two configurations can be added to form a whole, this configuration should be replaced with a monolithic configuration with the same content. In analogy to this, adding Ω-units should result in replacement of the smaller Ω-units with a single super-unit with same capability.

## 2.3 On Convertibility of Frameworks

In the most generic sense a configuration is a description of a system's setup. However,

such descriptions can be using vastly different formalisms. In the following sub-chapters I will describe major configuration styles I see being used in practice.

### 2.3.1 Functions

The probably most primitive example of what a configuration can be is a tuple of polynomial coefficients. Let's say, the behavior of the system is merely defined through a very static policy $f$ such as:

$$f(x) := c_n \cdot x^n + c_{n-1} \cdot x^{n-1} + \ldots + c_1 \cdot x^1 + c_0$$

*Equation 1: A polynomial function with configuring coefficients $c_i$*

In that case the tuple of coefficients $(c_n, c_{n-1}, \ldots, c_1, c_0)$ can be said to be the configuration. For functions relying on vectors or matrices of coefficients it is easy to expand this model to use vectors of matrices.

Similar approach is pursued with an "init file" which is basically a collection of defined key value pairs, as long as they are independent from each other. Otherwise such configurations are better explained with graph maps or programmed configurations.

Characteristic for such models is that none of the values is optional. Each value must be set explicitly or implicitly with a default. The size of the configuration must exactly match the number of degrees of freedom in a policy.

### 2.3.2 Spaces

A more sophisticated version of tuples are multidimensional matrices where each cell is representative of a position in a more or less concrete space. In concrete cases this could be a grid patch on a geophysical map. In more abstract cases it could be something like the "position in the table of elements". In reconfigurable robotics, ones in which a set of identical components is forming the robot's body, such spaces are called *lattices* [23][12] and are used to locate the position of components in the body. Space formalisms intend to model absolute reference.

In such models, the identifier is placed exactly into the cell that shall "hold" the "object". The model does not prevent illegal configurations. Errors in configurations could yield objects existing in multiple places at once.

Yet another version of spaces is allocating certain properties to each space location instead of an object id (which would serve as a reference for looking up the object's property). In that case an anonymous object inhabits each cell of the space. Example of such spaces are vector fields used to describe physical quantities. In that case a *spaces-style* configuration is approximating a *functions-style* configuration.

Characteristic for space formalisms is that they can contain a special value *null, void* or *empty* and, commonly, space matrices are rather sparse and high-dimensional.

### 2.3.3 Graph Maps

In systems where objects can be connected without special interfaces, for example because they do not model flows, a system can be understood as a plain graph. Such plain graph consists of arcs and nodes. Arcs connect the nodes. Different types of arcs can exist for different types of semantics. Nodes can contain special attributes. The goal of this description formalism is to highlight the relationship among the parts in a configuration.

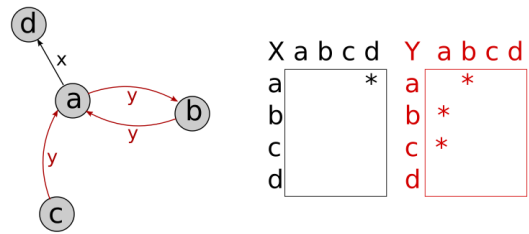

*Figure 4: A graph can be represented as matrices which represent relationships of type x and y between properties a, b, c and d. The * represent the existence of a directed relation.*

A very straight forward description of the configuration in a *graph* is made of a series of boolean connection matrices (as shown in figure 4) and of a collection of (typed) attributes. Each attribute serves as a node in the graph. It is the structure of the graph defining the "objects" and "object properties". If there are $i$ attributes and $j$ semantically understood connection types, then the total configuration is of size $i^2 j$ for two-way connectivity – but highly sparse. Since very sparse matrices are preferably stored as explicit assignments between node IDs, it is relatively easy to ad-dress the Boolean values by a tuple of kind ($ID_{from}$,$ID_{to}$). Because the IDs can be used as column and row indexes, even in very large connection matrices it is relatively easy to maintain stable dimension ordering which is an important feature in order to define operations on and comparisons of configurations.

### 2.3.4 Programmatic Configurations

Some products are using programmatic configurations for their setup, for example in form of Lua, JavaScript or Python scripts.

The advantages of such configurations are threefold:

a) It is possible to create very large but very sparse configuration descriptions in memory.

b) It is possible to create conditional variations of a configurations depending on environmental conditions.

c) Configuration is compressed: A pattern in configuration can be expressed as few algorithmic rules, like for example "every second connection is active".

If we accept the temporal nature of the executed configuration program, then it can be understood as a mechanism of reconfiguration along certain configuration space dimensions. This process is shown in figure 5.

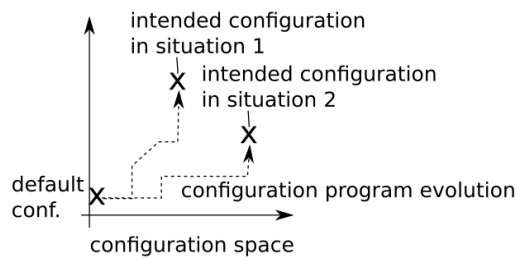

*Figure 5: A configuration program is executed in order to move a configuration from a default initial state into a context-aware final state.*

Despite that the process executing the configuration algorithm is evolving over time, in terms of the virtual time the process is "spontaneous" or "immediate". The configured system never interacts with its environment during this process. The difference between *real* and *virtual* configuration time is an important idea that I will expand on a bit more later.

### 2.3.5 Transport-style Configurations

Transport-style configurations are expressed in terms of objects with locations ("objects" hold "locations") and the configuration space is totally implicit.

Adding objects also adds a tuple of coordinates to the configuration space. Depending how strongly the objects interact, there is more or less independence between their partial configuration spaces (cf. figure 6). Connected objects result in *entangled spaces*. For example, if dim0 and dim1 represent position (x,y) of object 0 and dim10 & dim11 represent position (x,y) of object 6 and if object 0 and object 6 are "connected" then the values in those dimensions cannot be chosen independently anymore. This is to some degree also true for elastic relationships where entanglement is elastic. Such elastic entanglements are often found as hysteresis relationships between dimensions.

Configurations in transport-style are very well suited for representing transport problems, hence the name.

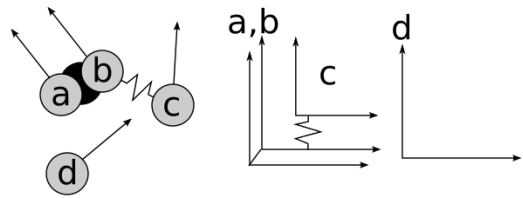

*Figure 6: In transport-style configurations, valid configurations are constrained by subspace dependencies which can be reasonably called "entanglement". In case of a and b, the subspaces are perfectly entangled. In case of a and c, the subspaces are elastically entangled. In case of a and d, the subspaces are independent. The observed entanglement can change over time: It can become independent, dependent or elastic depending on the models of moving elements' behaviors.*

### 2.3.6 Accumulation-style Configurations

Accumulation-style configurations are a mixed variant of *spaces* and *functional arguments*. Typical applications making use of accumulation-style configurations are found in physics where the state of a system is expressed in terms of a multitude of quantities. Such quantities are often governed by differential equations and corresponding spaces are called *state spaces*. What is similar to *function-style* configurations is that the values are not optional but what is different is that not all arguments can be feasibly overwritten (because of functioning as internal states) – only partial configuration updates are practically possible. What is similar to *space-style* configurations is that the quantities relate to more or less abstract places. A pressure value of a tank is describing a common property in a certain area of a plant. What is different from spaces is the idea that accumulators and their governing equations do not represent the equivalent of full meshes of space. The connectivity in accumulation-style configuration can be considered as a degenerated mesh that can only support transports also supported by some reference system. However, there are examples in which full meshes are used: in gas and fluid simulations we find full meshes which are governed all by the same equations (e.g. Navier-Stokes equations).

Because accumulation-style configurations are often used to represent geographically allocated containers, conversion between transport-style configurations and accumulation-style configurations is relatively easy (cf. figure 7). It is more the mathematical convenience defining which model paradigm should be taken.

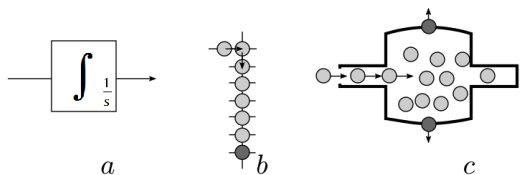

*Figure 7: A key element of accumulative models is the value in the integrator (a) and its scaling coefficients. Those can be translated into abstract (b) and realistic (c) transport models and vice versa.*

In the process of converting transport problems into accumulation problems information about many positions is lost but since such conversions are associated with problems of high granularity the exact positions are often irrelevant. Where necessary, positions can be restored by generating them randomly to the necessary mass effect.

### 2.3.7 Interface-based Configurations

In telecommunication disciplines many problems are mainly characterized in terms of which interfaces are provided, the exact definition of the interface, the interface grouping (reuse of interfaces) and joint interface constraints (all kinds of performance parameters).

More precisely. these configurations consist of mainly two parts:

1. Routing policies (cf. figure 8)
2. Interface parameters (cf. figure 9)

The routing policies as shown in figure 8 are rather high level and require a transformation step before being usable. That's because communication nodes often do not have means to fulfill their mission technically on this level of abstraction. For example, a computer with Ethernet interfaces must have means to deduce a sending policy for his Ethernet adapters when given only the abstract IP addresses that (except for few cases of self-reference) cannot be directly communicated with.

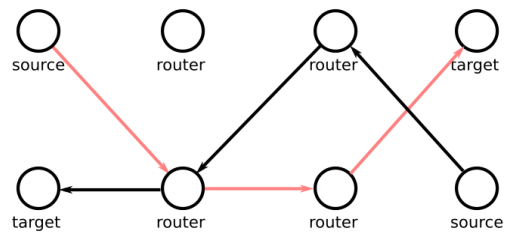

*Figure 8: In communications a configuration consists of routing policies which implicitly realize an optimal transport plan between sources and targets. An explicit optimal transport plan in form of a graph (as shown above) is a theoretical tool to understand a problem but is often practically infeasible as communications between sources and targets are far too many.*

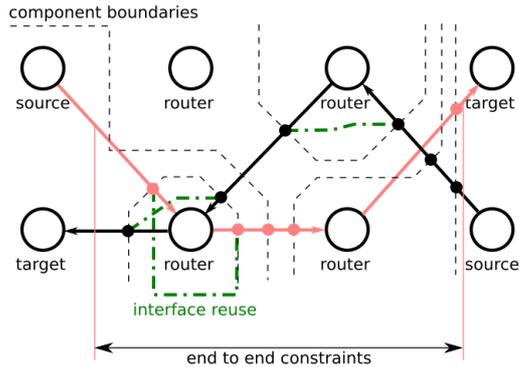
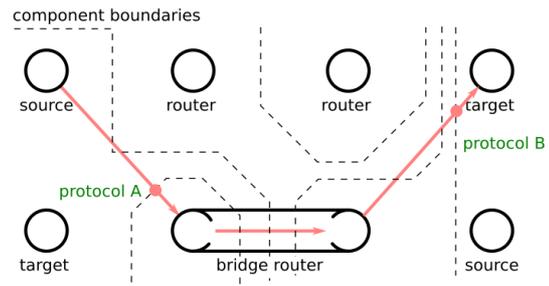

*Figure 9: A communications configuration also consists of node allocations and interface declarations on component boundaries hosting the communication nodes. Satisfaction of end-to-end constraints is key activity when designing distributed real-time applications. Optimization of bulk performance is influenced by the type and configuration of interfaces on communication paths. Each blob (red, black) represents an interface. Green lines indicate interface reuse and hence a shared constraints impact.*

This deduction is performed by assigning abstract communication nodes (or units) to component boundaries. Component boundaries are nested because component are nested. Some components are responsible for wrapping the communication in additional transport frames. For realistic systems, this leads to creation of eventually many interfaces (cf. figure 9) which all require information that must be provided before becoming operational. In order to reduce the amount of information necessary to configure a communications system, some of the degrees of freedom are restrained by interface reuse. In that case multiple theoretical communication paths are crossing the same interface. Practically, this aggregates the sharpest constraints on them from all communication paths.

Some of the communication nodes can be allocated to more than one component which is actually very common to do (bridges, gateways). In such cases the number of interfaces to be used can be eventually vastly reduced (cf. figure 10). Because the exact choice of interface and reuse of communications between them, communication configurations are highly discontinuous and non-linear making them brittle. Creation of robust and efficient communication networking solutions is a complete field of study [24].

*Figure 10: Same communication nodes can be allocated to different components (multi-allocation).*

For the scope of this paper it is enough to conclude that communication network configurations can be expressed as a graph where each communication node and interface can have properties and certain types of arcs for connecting them (holding information about direction, link speed, protocol type, etc.). This only extends to the configuration model shown section 2.3.3 in the way that nodes cannot be arbitrarily connected and that some minimum complexity is to be expected for such models. Otherwise they still consist of link matrices and value/property lists. Information, such as allocation of nodes to components is helpful for verifying correctness of new configurations but are otherwise redundant pieces of information.

Components can become explicit in such graphs when modeled as nodes. Allocation between communication nodes and component nodes will require a dedicated type of link. However, that still does not change the technical concept of how the configuration is represented formally as matrices and lists and this is a key feature for enabling convertibility into configuration spaces.

### 2.3.8 Configuration Strings

Many configurations are provided as strings of characters. This model is very useful when configurations are made by human editors who want to use very basic input tools. Configuration strings follow a certain syntax and grammar (temporal structure). Parsers and interpreters are used for extracting useful configuration information for machines.

In terms of representation there are several layers of information stored in a string which are more or less obvious. What is meant by that? Strings can be represented as a point in space where the space has *n* dimensions and *n* corresponds to the length of the string. However, the patterns represented in this way are often not easy to process. For example:

"property=value"

"property = value"

" propertY =    VALUE"

will look very different in their basic string representation and this makes the use of such a representation very difficult. Much easier to process is a distinct dimension *property* with an atomic value *value*.

However, as already remarked, strings eventually contain more than a single atomic value to translate. For example:

> "a a b c b b a c c" could contain the following semantic content:
>
> Level 0 interpretation:
>
> a, a, b, c, b, b, a, c, c
>     ($\rightarrow$ 9 basic dimensions;
> unfortunately, it is how computers see things)
>
> Level 1 interpretation:
>
> phi, psi, rho ($\rightarrow$ 3 meta dimensions)
>
> "a a b" : phi
>
> "c b b" : psi
>
> "a c c" : rho
>
>
> Level 2:
>
> G, H ($\rightarrow$ 2 meta-meta dimensions)
>
> G : phi, psi
>
> H : rho + TERMINAL
>
>
> Level 3:
>
> A : G, H ($\rightarrow$ 1 super dimension, possibly what the human reader sees in the string)

It might depend on the application to define which levels to extract, i.e. one, some or all possible "contents" from the string and to incorporate them in the configuration description. As result of such choices, two string-based configurations can appear near or afar depending on chosen level of interpretation. This level is normally identified by the appearance of invariant contents.

In figure 11 a low level interpretation of configuration data is presented. The space used to define configurations is functionally ambiguous. Several configurations presented as input will behave exactly the same on the plant. This is a signal to the designer that the input format for a configuration is not well chosen: Optimizing the plant performance is only done by defining movement between isolines. For this to work properly, domain-specific knowledge is often required by the configuration mechanism.

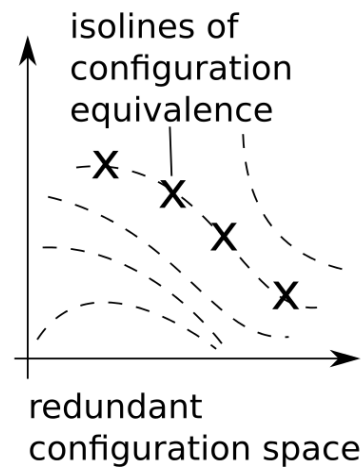

*Figure 11: If the dimensions used for configurations are not atomic (like in strings) then this can result in redundant configurations subspaces where multiple atomic configurations result in same result.*

### 2.3.9 Input and Output Configurations

The problem shown in figure 11 is related to the difference between *input configurations* and *output configurations*. Input configurations go into the configurator and output configurations leave the configurator. In some cases the mapping between the two can vastly misguide about the effects. Figure 12 illustrates the problem: Configuration space for direct input can look vastly different from the "effective" configuration space (function space) which is responsible for producing the function or structure of interest. Configurations which are near in input configuration space can be far away in the function space. This makes systems using such configuration spaces tendentiously brittle.

Deterministic mapping of configurations between input and output space does not prevent surprises: Paths taken to attain a new configuration in the input space need not reflect anything similar in the output space. In figure 12 we see a short and a long path between two configurations. A system that is trying to interpolate between the two configurations (start and end) in a fast and linear way might be quite surprised what it did in result. Reconfiguration performed in the representative input space must sometimes choose a "quality path" between two configurations which is not necessarily linear or easy to obtain analytically.

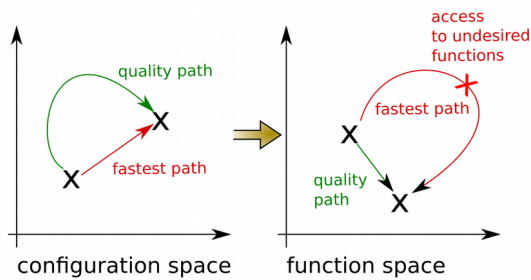

*Figure 12: The exact relationship between the space used to configure the system and the space that actually represents the system plant may have intriguing effects on reliability, quality and speed of reconfigurable system.*

### 2.3.10 Understanding Configurations as Patterns

Previous sections discussed popular problems which are dealing with configurations of some kind. The discussion of them had the purpose to demonstrate that the different types of configuration can be often boiled down to some very basic concepts like a matrix, a tuple or lists. Depending on the exact nature of the problem, sizes of these basic objects may vary. Going from one legal configuration to another legal configuration may increase the amount of bits of information that is necessary to describe it.

Even if the domain-specific configurations are best suited to represent the problem in their domain and to perform optimizations on them, I don't see any real obstacles for transforming those problems into a fundamentally uniform formalism, that of a *point in a configuration space* (cf. figure 13) because it is not needed to define the exact number of dimensions for this space.

This approach is frequently chosen for machine learning purposes. For example, in theory of support vector machines the dimensionality of the segregation space is assumed "as large as necessary" and "as compact as possible". However, in machine learning we need to guarantee that dimensions never change their meanings as further samples are presented.

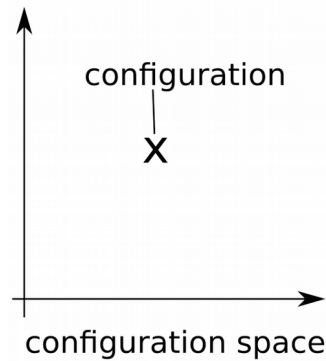

*Figure 13: A configuration is a "pattern" - a point in configuration space.*

Guaranteeing such stability for all configuration formalism could prove very difficult to achieve. However, for understanding the activities related to (self-)configuration the actual attainment of such stability in practice is of secondary priority. Such practical concerns become more pressing if we wanted, for example, to measure distances between configurations or if we wanted to classify legal configurations from illegal ones: We would run into various problems such as achieving practical dimension stability or severe space fragmentation.

### 2.3.11 Configuration Space Fragmentation From Dimension Representation Decisions

Each dimension can demonstrate its own level of granularity. Some dimensions can provide values from an open set (like for example real numbers) or from a closed set (Boolean values). The granularity of dimensions and their openness is often not as much of a problem but the idea of a dimension also quietly assumes an ordering of values which would allow "movement" along the dimension. This is of course a problem because such movements cannot be defined for unordered sets. A way to solve this is to decompose dimensions into binary dimensions and let a superimposed mechanism detect and validate the xor relationship among them.

*Figure 14: Configuration properties which are not naturally having an order can be either explicitly arranged (then never rearranged) or decomposed into (almost) independent dimensions. A validator must verify the legality of resulting configuration.*

Link maps for graphs naturally obey this approach as long as they contain only binary values (link exists or exists not). In case that these maps contain IDs to link objects, some transformation is necessary in order to prevent interpretation of the ID value as a configuration parameter (because it is not).

However, if the links contain values like for example *weights* then of course those are configuration parameters. Multi-valued links could be decomposed into several single-valued links.

## 2.4 Identification of Legal Configuration Subspace L

The conversion of various configuration types into a configuration space formalism is very convenient but in all realistic scenarios there exist some constraints regarding feasibility of configurations in such space.

The default configuration space model assumes that all dimensions are independent and unconstrained (fig. 15).

*Figure 15: A benign space consists of many continuous valued dimensions with no hard constraints on configuration legality: All space comprises the feasible configuration space L.*

For any realistic scenario such benign spaces are the exception. More frequently, spaces are divided up into areas with feasible configurations (L-space) and illegal configurations (N-space) (cf. figures 14 and 16). Some of the constraints are introduced during a configuration model conversion (fig. 14) and others are sourced in the domain. For example, a configuration is infeasible if it configures more power consumption than power production.

*Figure 16: A configuration bridge in configuration space.*

The subspace of feasible space in the configuration space can be arbitrarily small. The amount and distribution of L-space is implicitly defining which methods can be used in order to find a new configuration.

Finding a feasible configuration requires an identification function which is generated along with the configuration space in the conversion process. How to compute such an identification function is a research area on its own. Cost functions and inequations are typi-cal models used for continuous spaces.

Readers interested in construction of identification functions for highly discrete and fragmented configuration spaces can relate to literature on configuration management where finding and validating feasibility of configuration is a key activity [25]–[27]. In configuration management the configuration space dimensions are modeled as feature trees with various dependencies. A feature selection validation step is the equivalent of the here required identification step but it is open how to get a "direction" towards legal configurations after bad configurations were detected.

In figure 16 the L-space is contiguous and covers source and target configurations. In that case the areas connecting the source and target configurations are *configuration bridges* by which a reconfiguration controller can choose to transit incrementally.

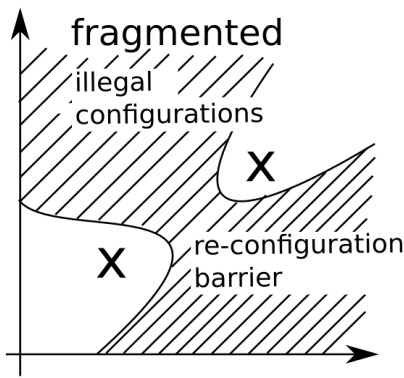

*Figure 17: In a fragmented configuration space there must be a single spontaneous reconfiguration step which can skip the complete configuration space barrier.*

There is no limit how severely the configuration space is covered with illegal configurations. In fact, feasible configurations can lie on configuration islands which have no feasible intermediate configurations. In that case configurations are isolated by *configuration barriers* – the exact opposite of configuration bridges (cf. figure 17).

Please note, that configuration barriers do not imply that a system can never reconfigure from one island to another. It just means that if the reconfiguration cannot jump the barrier given available bandwidths then it will fail. This would be the case for slow configurators. Chapter 7.2 expands on this problem.

# 3 The Configurable Plant

The configurable plant is the essence of a system with self-configuration capability. It contains a set of elements which can be rearranged or parametrized in order to tune its function.

The exact implementation of a plant can look vastly differently from plant to plant, so that we cannot investigate plants on grounds of a particular formal model. However, what can be done is some general considerations which do not rely on implementation specifics. One such consideration is the assessment of a worst case and a best case.

## 3.1 Static Plants

Let us consider a static plant at first. A static plant will not generate its own behavior. In worst case, such a plant will implement a dedicated sub-plant for each possible configuration in its configuration space, as shown in figure 18. A sub-plant is simply some kind of function taking the plant input as input. In theory, all computations in the plant occur simultaneously. The multiplex selector decides which output is forwarded to be the plant's output.

The sub-plant functions discussed here are hard coded ("atomic sub-plant"). Hence, any sub-plant is taking only the functional input as argument. The reconfiguration is simply a multiplex selector operation. If it is not, because a sub-plant is taking configuration arguments, then it can be always decomposed into less flexible sub-plants. Involved variables are removed from the sub-plant and then incorporated into the legal space of configuration input of the major multiplexer.

Consequently, the true size of plant configurations (I will refer to this by the letter $\Psi$) will only depend on the size and resolution of the most compact configuration space required to represent the necessary variability of a system's plant. It is exactly this amount of information necessary to safely identify a singular sub-plant with a multiplexer as is shown in figure 18.

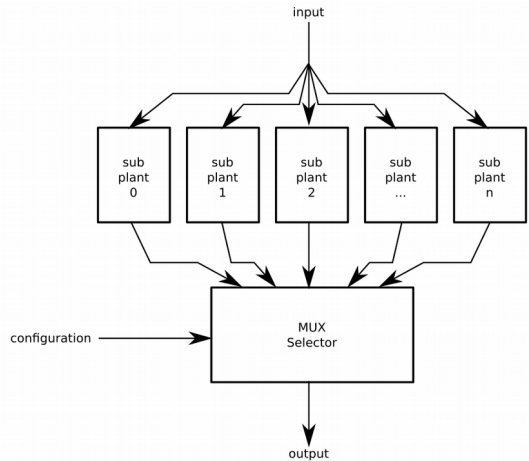

*Figure 18: At worst, a static plant will implement an individual mechanism for each configuration in the configuration space.*

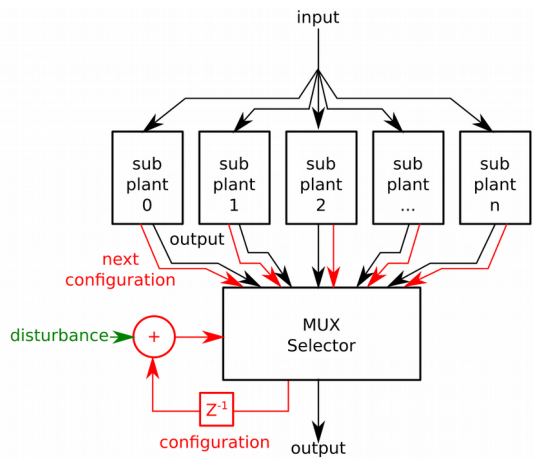

*Figure 19: A dynamic plant is generating its own configuration input. External configuration input can be permanent (addressing) or if it is volatile then it can be understood as some type of disturbance to such plant.*

### 3.2 Dynamic Plants

Dynamic plants are same as static plants but will follow a particular configuration mapping between configurations. For that purpose any atomic sub-plant also generates the address of the next configuration to be used (cf. figure 19). If the address is used with no alteration then plant would be autonomous and hence not reconfigurable. In order to become reconfigurable, the plant must surrender its autonomy to a disturbing source (which is the *configurator* or *controller*).

However, a disturbance can have many factors in it. It could contain information from a controller but could also contain random noise in order to model unreliability of sub-plant activation.

### 3.3 Sub-Plant Configuration Space

The configuration input to the multiplex selector is normally not just a sequence of bits. This offers an opportunity to visualize the processes in the plant a little bit better. Let us assume that a plant would vary by two distinct criteria – yielding a 2D configuration space (cf. figure 20). Each configuration is then identified by a coordinate $i$ (vector or tuple of integers or mixed). For sake of imagination let us assume $i$ was two constants for a polynomial function used to implement each sub-plant. Of course, a real implementation would not implement myriads of plants but only a single function with proper arguments. However, having just more functions with variable parameters is not going to clearly help us understand the role of reconfigurations caused by the controller.

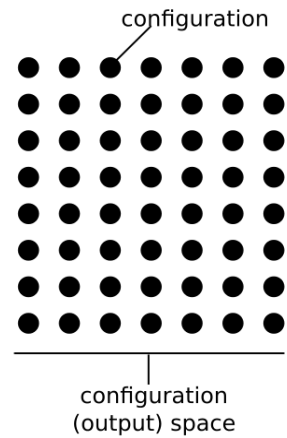

*Figure 20: A configuration space of a plant consists of "collection of points". Their relationship is not necessarily clear but most models assume independent dimensions and gird-like ordering. This assumption is a quite strong input to the model which is not required. The grid is just helpful to understand processes in a configurable plant, visually.*

I chose to depict the reconfiguration processes in a discrete manner. This approach is suitable for configuration spaces where discrete dimensions and continuous dimensions are mixed (where the configurations have really many close by configurations). There is always an "active configuration" in this space which forwards its activation to the next configuration. Depending on the nature of the plant there can be further activations – for example if there are two parallel running plants with exactly the same configuration space.

A configuration is at minimum a constant function delivering a multi-valued entity ($f_i$) which is only meaningful to the outside world. In more complex cases it is a complex static function. It is important to note that in this model the input (black lines in figure 19) are never used to compute the transitions relationship between configurations $j_i$ (red lines in figure 19) - a clean pass through:

$$c_i := j_i, f_i(input)$$
$$c_i := j_i, i$$

*Equation 2: The target configuration address $j_i$ never depends on input from environment. Interaction with the plant is solely understood in terms of disturbances. In a degenerated special case, the function f can be defined to be the constant configuration parameter i.*

Also, I do not consider dynamic or configurable sub-plants because this results in a "plant in plant" model which can be decomposed into a flat plant model as was shown in figure 19. Why this can be done is shown in figure 21: If the nested plant is a dynamic component with an own multiplex selector to organize its function then if the boundaries of the sub-plant are removed then the multiplexers form a direct cascade. Such cascades can be easily removed with a monolithic multiplex selector which is offering input ports for the (so far) nested sub-sub-plants. Practically all that is necessary to achieve this is to re-compute addresses provided by *all* sub-plants – which is also the answer to the question why such sub-plants could be grouped: In order to avoid "treatment" of address ranges which are not of concern to the grouped sub-sub-plants.

Let us look at a configuration in the output configuration space in context of a dynamic plant: Each configuration is explicitly identifiable and can be addressed as target by as many sub-plants as there are but it will deliver only one reference to a successor. It is possible to define alternative successors (figure 22) which could be selected by an additional parameter in order to model system unreliability (this concept can be extended in order to attain first order Markov models).

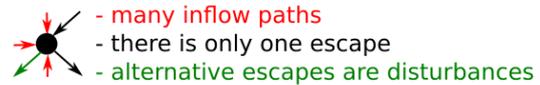

*Figure 22: Legend for following diagrams.*

Let us assume that most configurations simply refer back to themselves. This would reduce the overall plant dynamics to a static plant. This setup can be narrowly altered by creating activation chains for particular configurations as seen in figure 23.

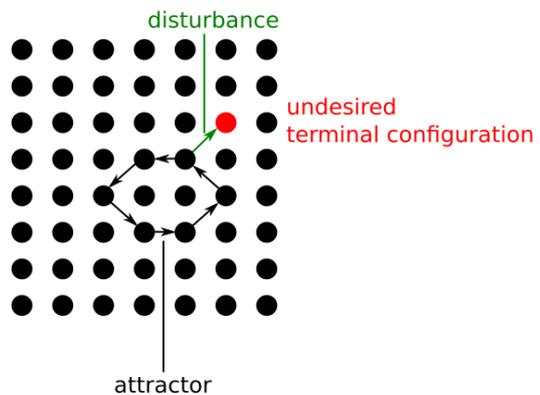

*Figure 23: An attractor that would be called a "program"*

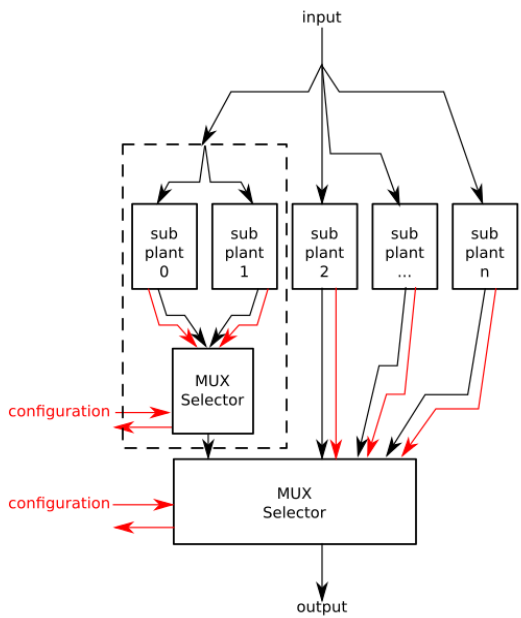

*Figure 21: A nested sub-plant can be simply expanded to a flat setting by transferring the function of the inset multiplexer to the final multiplexer.*

If the plant is set to one of the "chained" configurations then it will keep on changing its states depending on its basic clock. In fact, the clock speed of the plant is a configuration parameter that is not part of the configuration space. A higher clock speed will make the system "compute" faster and lowering the clock speed will make the system "compute" slower. For practical reasons, where systems need to model transition strategies[2] of varying speeds, systems are advised to implement those speed changes by offering transitions stretching various ranges between the configurations.

The attractor in figure 23 is normally not called an attractor because it is surrounded by a (close to infinite) number of self-targeting configurations (point attractors). More frequently, we find the term *program* to be used to describe this situation. If there is an error then the program is interrupted by going into an *undesired terminal configuration*. Technically, the system plant arrived at one of its point attractors but colloquially we would say that the program has hung itself up or that it terminated.

In order to improve resilience of the plant it would be good to increase its tolerance toward activation of configurations outside the main paths. This will demand that surrounding and not really involved configurations will act as return guides toward the actual program. This surrounding corrective addressing between the configurations makes the attractor really stand out as an attractor in traditional sense. With a dense corrective addressing field any false jump in the configuration space (a "disturbance") is going to be corrected by the plant (cf. figure 24).

So far, the plant is performing its action spontaneously and without external intervention. However, the plant performs exactly one unconditional policy. Since the plant model has been defined in terms of sub-plants representing configurations according to eq. 2, the plant is not capable to perform conditional behavior alterations.

By the here proposed model all conditionally working plants must be decomposed into more fundamental units which lack conditionality. The conditional part shall become subject to controller-plant interactions after decomposition. Nevertheless, "decisions" are still possible by disturbance: A layout of configuration space exploiting systematic type of disturbances could suggest making decisions based on input. However, the attractors in the process have not changed (cf. figure 25).

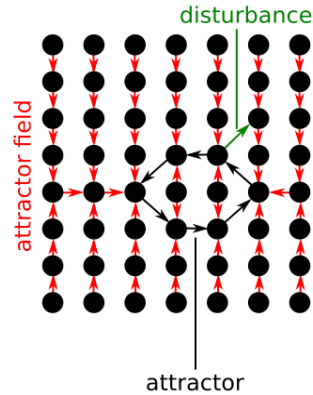

*Figure 24: An attractor field can make plant programs resilient to disturbances.*

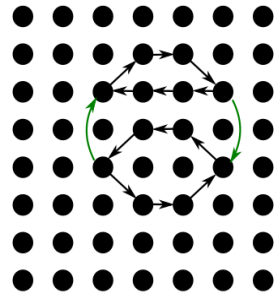

*Figure 25: Relying on systematic disturbances can make the plant appear as if it did perform "decisions" but the policy has been static all along.*

Since reconfiguration can be understood as explicit decision making, the plant model relies on the reconfiguration controller to perform decisions regarding the flow.

In figure 26 we see an example of conditional plant policies and two decision points A and B. Those can be understood as two incoming configuration parameters which can be provided externally via the multiplex selector as was proposed in figure 18. Since A and B are defining binary choices, the most compact configuration representation $\Psi$ is 2 bits. Since bit for B is only relevant for a certain A value the average $\Phi$ representation in storage will be 1.5 bits long and 2 bits max.

---

2  or "policies"

Selecting between the policies can be understood as introducing additional dimensions and the changing of A and B values as dimensional transitions. In that case the configuration space would look like a higher dimensional configuration space with planes.

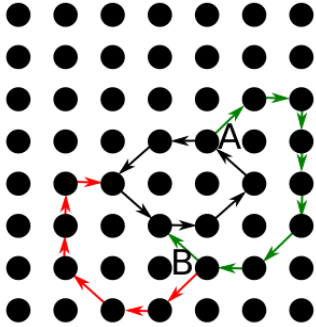

*Figure 26: Three policies connected via two points of divergence A & B.*

For the example case this is similar to making the configuration space a 4D vector field. However, vector fields imply a Cartesian relationship between the configurations which is not required to be true for the models of reconfigurable plant to work. In fact, for sophisticated systems a vector field would be rather an exception[3].

The configuration process can be better explained as reorganization of mappings between the configurations:

$$c_i := j_i(A, B), f_i(input)$$

*Equation 3: Remapping of plant given our example configuration dimensions A and B*

In figures 27-30 we can see the resulting programs / attractors which can be obtained from configuring values for A and B.

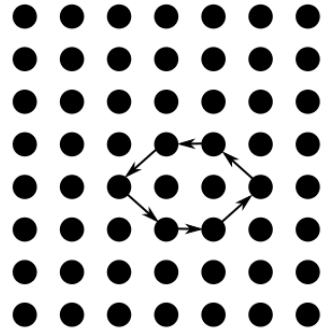

*Figure 27: A=0, B=0*

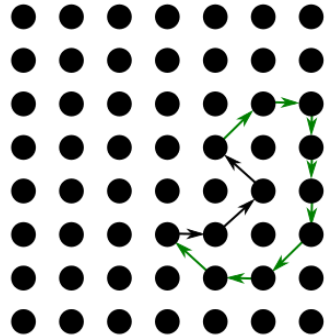

*Figure 28: A=1, B=0*

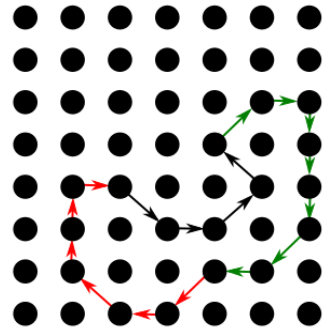

*Figure 29: A=1, B=1*

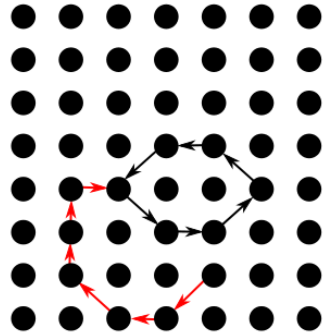

*Figure 30: A=0, B=1*

---

3   The reader be warned that there is a difference between vector fields and configuration spaces, e.g. no affinity holds true in configuration spaces and configurations spaces do not allow overlapping movement of plant and controller unless the configuration space has been prepared to act like a vector field. However, it is out of scope to discuss the differences here. It is something for future work to explain in detail.

### 3.4 On Addresses Computed from Input

I would like to briefly comment on why a sub-plant does not make its new target depend on its input. It could be possible to define a configuration like this:

$$c_i := j_i(input), f_i(input)$$

*Equation 4: Conditional plant configuration*

The succeeding address $j_i$ depends on the input values to the system. The downside of this model lies in the way how we can understand decision making in the self-configuring system: Using eq. 4, the environment, the plant and the controller are all involved in "decision making" by having a parametric function. I find it not clarifying to have a set of functions depending on magical properties of the input. The function $j_i()$ needs to classify input in order to perform a decision. If timely decisions are required then $j_i()$ can become dependent on past values of itself or past values of input (input sequences). Alternatively, the configuration parameter $i$ can be made dependent on input or input sequences. However, all this flexibility only introduces new questions on how these functions and dynamic relationships fit into picture of the here sketched interactions between plant, controller and configuration storage. With a static $j_i$ we can rely on simple transformation arguments made of decompositions and relative addressing of configurations, on information processing principles and optimization theory. A good systemic self-configuration theory should propose mostly a unique place for each "feature" to be explained.

### 3.5 Summary of Chapter

In this chapter an abstract "worst case plant" has been presented for which several considerations about its function can be made but independently of a particular formal or technical implementation. Decision making in such plant is mainly performed by external reconfiguration but a plant can be designed to exploit systematic disturbances in order to switch between attractors (pseudo decisions). Reliability of plants can be improved by introducing corrective fields.

## 4 The Configuring Controller

The purpose of the configuring controller is to observe the plant, the plant's performance and to execute plant reconfigurations. The controller makes tactical decisions to varying degrees of sophistication. It estimates the benefits and costs of plant modes (economic assessments) and minimizes risks and costs related to mode transitions.

### 4.1 Transitions

Since self-configuration and reconfiguration capabilities are demanded for systems which are exposed to changing loads, goals or other conditions, it is important to assess changes (or their absence) in terms of how they will affect a dynamic system.

An important aspect in that assessment is the relationship between *real time* and *virtual time*. Real time is of very high importance to a system's operational success. In the specific area of autonomous, self-maintaining systems, synchronization between internal and external events is of utmost importance to system's chances of survival. Failure to respond in time can pose a significant risk for a system.

I guess, the problem is best known to most people as the "IT is down" story: Changes have to be made to the configuration of servers and networks and this cannot be done in hot manner because chances that people will sit around are very high and very costly. The IT staff takes down the system during the night or during the weekend in order to minimize experienced failures of service and hence risks of costs.

In order to understand the risks coming from reconfiguration an example is showing the relationship between real time and virtual time in figure 31. In that example the system's plant is represented by a red dot. The system plant can jump between three modes (three configurations). The system interacts rapidly with the environment in each of the modes. However, even then the responsiveness of the system varies. For example, in mode 1 the system can be performing the sense-plan-act cycles at a mediocre speed but in mode 2 it can greatly improve its response time (this could be the reason why the red system has different modes in the first place).

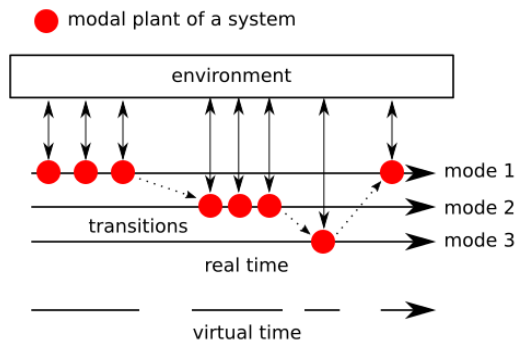

*Figure 31: Reconfiguration of system's plant can temporarily disable it for a transitional period of time. During that time a system is not responding to its environment. Even if a reconfiguration consumes long time to be performed, a system perceives the change as spontaneous on the inside.*

Nevertheless, changing between modes costs the system transition times which are not related to the speeds obtained during modes. Those are the times required for the "rebuild" of the system's internal configuration.

During the time of *rebuild* or *reconfiguration* the system is not effectively progressing – its virtual time is standing still. A reconfiguration in virtual time is instantaneous or *spontaneous*; a change in one unit of operation (let's say "cycle").

In mode 2 the virtual time is progressing faster than in mode number 1. Therefore, there is no clear relationship between virtual time scales in the different phases of the system and system's real-time.

If the time necessary to transit from one mode to another exceeds system's constraints, like for example the maximum time to respond or the amount of necessary energy to perform it, the system could implement a strategy in which the configuration is approached iteratively (cf. figure 32). This gives raise to two unique kinds of configuration mechanisms: spontaneous and iteratively reactive mechanisms.

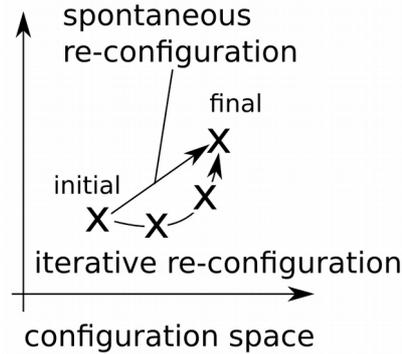

*Figure 32: Spontaneous vs. iterative / reactive reconfiguration process.*

Figure 33 shows the advantage of iterative, incrementally adjusted configurations: The system can finish smaller adjustments earlier and hence respond to events in the environment while still pursuing greater changes in its configuration.

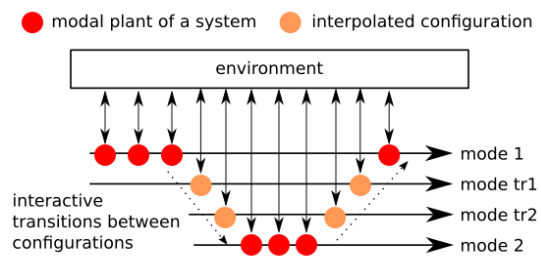

*Figure 33: Iterative & reactive reconfiguration process. tr1, tr2: transitional configurations: The supposed intermediary modes tr1 and tr2 need not be structurally similar but similar by effect.*

Optimization of transition risks and costs will require some basic pieces of information:

- How much change needs to be made between the IS configuration and the TARGET configuration? (the delta)

- Given a set of alternative implementation strategies, which strategies show the least costs for performing the delta?

The controller can rely on game-theoretic approaches or approaches known from dynamic programming. For example it could compute the maximum damage over an evolving tree of real-time activities (cf. figure 34). A damage is the amount of constraints violation in a given state. States follow states and states can lead to several other states which will yield damages. Computing damages over the discourse of system evolution is an important idea in order to assess the maximum amount of permissible reconfiguration time $t_{reconf}$. Estimation of this variable is key parameter in deciding for or against a given reconfiguration strategy α.

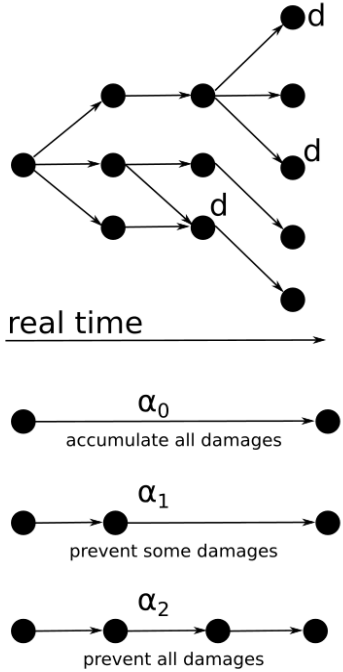

*Figure 34: Visualization of what equations 5 and 6 mean: The more damages are accepted by c the more monolithic reconfiguration can occur.*

$$D(\text{A}) = \max_{\alpha \in \text{A}} \sum_{t=1}^{t_{reconf}} d(t, \alpha)$$

*Equation 5: Compute maximum damage d for any reconfiguration strategy α for a real-time window of $t_{reconf}$ width*

This will require look-ahead simulation (or deep searches). Equation 5 is similar to the Bellman equation but with no elements of discount and not expressed in terms of rewards but of damages. D(A) is the damage for a given strategy. Reducing the $t_{reconf}$ can greatly reduce expected damages.

Only after identifying feasible strategies with acceptable damages, the controller should employ regularization terms in order to choose a strategy of least effort. A simple damage limiting argument *c* can help to drop any strategy from the set of eligible strategies $\mathring{\text{A}}$:

$$\mathring{\text{A}}(\text{A}, c) := \forall \sum_{t=1}^{t_{reconf}} d(t, \alpha) < c \; \exists \alpha \in \mathring{\text{A}}$$

*Equation 6: $\mathring{\text{A}}$ represents a collection of transition strategies eligible for cost-based selection and c is the maximum tolerated damage of strategy*

It is important to relate the cost of transition strategies to something meaningful in context of the real system environment. Cost could be the amount of floating point operations needed to perform a reimplementation of plant. However, time should never be an element of the cost factors because it is already implicitly contained in the summation of costs along the reconfiguration strategies. For example, a single reconfiguration step requiring 1 MFlop has equal cost to a strategy running 10 x 0.1 MFlop. In real time both strategies could run in 1s of time. It is not naturally given to prefer a single step reconfiguration. Frankly, multi-step reconfigurations are associated with higher overheads and hence costs. Given a collection of small risk strategies, shorter (in terms of real time) strategies will be preferred based on their costs. However, this is only an experience-based statement. It is easily imagined that reconfiguration costs sink dramatically when they are stretched in time. In analogy to a car that is experiencing air drag, moving a little bit slower implies significant gains on mileage. A rational driver will therefore reduce vehicle speed until his strategy hits hard constraints ("arrives too late") or power consumption[4] goes up again because of inefficiencies. Because the inefficiencies and constraints change over time, a driver adapts his vehicle's velocity dynamically but neither very slow nor very fast speeds dominate.

The same will be true for any properly designed configuration controller. It will choose very fast or very slow reconfiguration strategies only in exceptional cases. In all other more realistic cases, the controller will prefer moderate reconfiguration speeds.

---

4 Please note that power consumption is that of the car and the driver. From this results a favoring of generally faster speeds.

## 4.2 Modes and Controller Optimization Techniques

In the here proposed model, the plant is only storing the configuration that it is in. Other configurations dwell in a repository (storage) until the controller selects them for on-plant implementation.

The controller must decide which mode to select. A controller can choose between explicit and implicit modes. Explicit modes are ones explicitly identified in the storage and implicit modes are configurations which can be somehow created from a possible space of configurations.

This brings two tasks to our attention: The first job is to obtain the most comprehensive set of modes to select from. The second job is to grade them and to pick the best solution out of them.

### 4.2.1 Deterministic Optimization Approaches

This class of configuration mechanisms is implementing deterministic, potentially provably optimal and correct optimization algorithms which reconfigure the system plant. This is particularly successful for problems expressed as continuous optimization challenges. In that case a system can perform techniques like simplex ascend or gradient descend in order to find the next optimal choice of values. Invoking the optimization under the same set of conditions will yield reproducibly the same configurations.

Not all of optimization techniques can be aborted earlier in order to deliver an intermediate configuration. This is often the case with algorithms which start with a default state (e.g. zero vectors) and which need to be running through a complete process before arriving at a result. Others, like error backpropagating algorithms, can adapt the configuration incrementally. However, such incremental improvements also rely on benign configuration representations. Unconstrained, continuous spaces are such benign conditions (cf. figure 15).

In figure 16 a much less benign situation is shown in which the configuration space is divided between areas with illegal configurations and feasible configurations. In this case there exists a configuration bridge between initial and desired configuration – this bridge is made of similar configurations which can be selected for transitional configurations.

Linear techniques rely on "hitting the bridge" in such case. For guidance, the reconfigurator estimates cost of a new configuration and tries to select new configurations with lowest costs within a range:

$$C(c_2, c_1) = a(c_2) + p(c_2) + q(c_2) + r(c_1, c_2)$$

*Equation 7: Guiding cost function*

$C$: Cost of new configuration $c_2$, given $c_1$

$a$: attractor field towards new configuration; decreases costs towards 0 for final configuration.

$p$: penalty field; adds very high costs when $c_2$ becomes illegal. The more $c_2$ violates constraints, the higher the costs.

$q$: quality field; adds moderate costs for configurations with potentially undesirable functional side effects, such as potential to oscillate.

$r$: radius field: adds costs to $c_2$ the further the new configuration is away from $c_1$.

In theory, the reconfigurator only needs to pick the configuration for $c_2$ which has the lowest cost in the total cost field $C$.

Applying backpropagation techniques to very high dimensional space of discrete and artificially ordered representations, maybe with sporadically valid configurations, is rather futile. In such scenarios configuration spaces are probably fragmented (cf. figure 17), i.e. no linear interpolations will yield reasonably valid configurations.

In such cases, paradigms based on more complex rule inference are often working better (classic symbolic AI) because they can "jump" in complex ways between feasible configuration spaces. Fragmented configuration spaces are not only full of barriers but the feasible configuration subspaces could be also very small (maybe even singularities in a sea of illegal configurations). In that case only sophisticated, knowledge-based subspace selec-tion techniques will yield reasonable new configurations.

### 4.2.2 *Approaches based on Randomness*

This class of configuration mechanisms is implementing educated randomness-based re-configurations. This type of method is very popular in models with great amount of non-linearities and discontinuities. Examples of such methods are simulated annealing or various forms of genetic algorithms. Stochastic algorithms with strong "education" can produce feasible configurations even in difficult spaces. Mechanisms based on randomness prevent that the system will repeatedly get stuck with the same error given certain impasses. On the other hand it also prevents reproduction of ideal system plant configurations.

### 4.2.3 *Approaches based on Memory*

A system could be lacking a proper mechanism for inferring better configurations when faced with insufficient performance. However, it could simply remember configurations which were sufficient (not necessarily optimal) and which it has found accidentally, before. The memory can be used in order to simply restore the memorized configuration and then let other limited, adaptive mechanisms optimize it. If newer, distinctly different configurations are detected then those could be added to memory together with triggers to activate its deployment.

In terms of the model shown in figure 38 this means that there is little difference between the input configuration space (storage→ configurator) and output configuration space (configurator → plant). However, I intentionally say *little*, as memorized patterns could have been recoded, compressed or extended with error correction data.

### 4.2.4 *Approaches based on Mixing*

This class of methods relies on mixing new configurations from several other sources.

*Interpolation*: This is very easy to do for configurations with relatively unconstrained and well ordered dimensions. Interpolation of new configurations can take 2 or more reference configurations. If these configurations mark the boundaries of feasible configuration space, creation of new configurations based on this approach offers little risk to generate a new configuration that is invalid.

*Extrapolation*: This process is a little more sophisticated than interpolation because likelihood that extrapolated configurations lie within constraints are lower. Extrapolation requires a direction and a range at which a configuration shall be extrapolated. This process can also rely on 2 or more reference configurations which are used to regress a configuration development function used for extrapolation.

*Assembly*: This approach select sections of a configuration which is then replaced with partial information from a reference configuration. This can be used to assemble / create qualitatively new configurations for which there has been no prior experience (cf. figure 35):

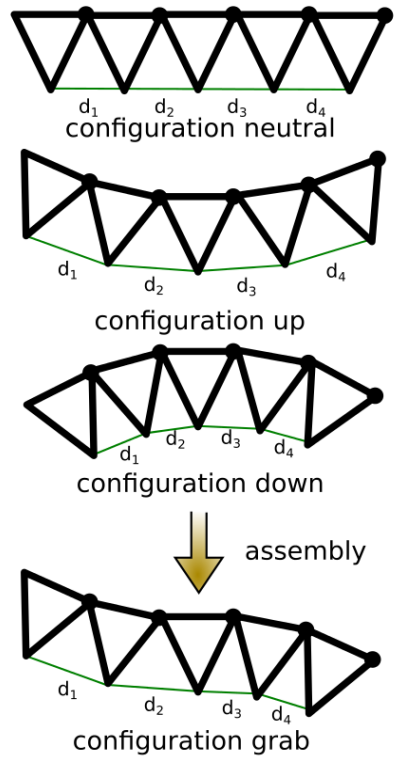

*Figure 35: Assembly of new configurations. $d_1$, $d_2$, $d_3$ and $d_4$ are configuration parameters.*

*Filtering*: If mixing of configurations is understood as a temporal process then it can also be understood as a signal filtering process. Applying filters (e.g. low pass) to a plant's configuration could slow down arrival at a goal configuration at the benefit of higher quality system behavior. It is also possible to combine certain "bands" from various configurators if the system supports more than just the one shown in the canonical model in figure 38. Normally, it is the role of the configurator to do the filtering. It could perform filtering in order to prioritize reconfiguration bits: Bits with largest impact could go first.

## 4.3 Synchronization of Speeds

Synchronicity can reveal itself by such activities as pushing the button at the right time, as balancing, swinging, doing the right bids on the market or regulating speed. Reconfiguration can play a role in system's ability to synchronize between plant and environment. In general sense, synchronization means a change to operation which improves temporal characteristics of the system. This can very well mean adaptation of delay, frequency or phase of plant functions, to tune or in fact even detune them.

In a synchronized state the interactions between plant and environment are without surprise, i.e. they yield no information to each other (cf. figure 36).

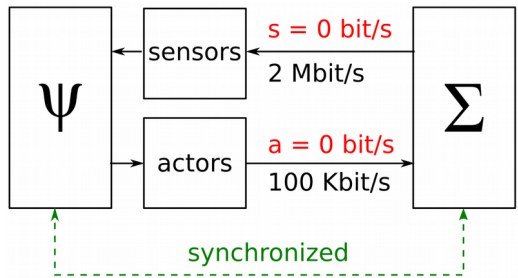

*Figure 36: Synchronicity of plant and environment.*

This kind of bidirectional synchronicity is a rather strong case and requires systems which monitor the interactions on both sides. However, most scenarios require only single-edged or single-sided synchronicity where the plant Ψ influences the environment system Σ with no deviation to its expectation. The amount of information exchange between the two systems can be measured using the Kullback-Leibler Divergence between predicted and measured data.

If a system relies on the fact that a particular configuration is arrived at then a system must initialize a program with just enough time ahead of the event in order to attain synchronicity. However, a dynamic, self-configuring system does not have a clear "starting point" because configuration activations are following the disturbance-successor schema as was described in section 3.2.

A system will face now two types of events by which a system can accelerate or decelerate its plant execution speed.

**On-track Disturbance**: The plant is either too fast or too slow in its operation. The environment has the resources to induce a disturbance which changes the activated configuration along the line of the plant program. This would be characteristic for mechanic plants with stiff coupling with the environment. If the configuration is reset back on the program path then the plant is too fast. If the plant activation skips configurations on the plant's program path then the plant is too slow. Detecting such events is necessary in order to increase or decrease the plant's clock parameters. Adjusted plant clock will reduce or eliminate unexpected configuration resets or configuration promotions. This type of synchronization does not require involvement of the configuration controller.

**Off-track Disturbance**: If the coupling between environment and plant is not stiff enough then disturbances resulting from bad synchronization do not result in disturbances along the program paths. In that case it is difficult to say whether the disturbance can be fixed by adapting the clock rates at all. In the worst case no disturbances occur at all because plant function is not directly influenced by the environment (e.g. plant is software).

Unfortunately, a "vector field" in plant's current configuration space is not able to capture the transitions necessary to always synchronize again. Either because no disturbance occurs or because the same configuration can be attained by a disturbance from various source configurations on the program path.

In that case the configuration controller has the duty to detect the deviation and to create a plant reconfiguration for the individual case in order to return to a synchronized state.

## 4.4 Error Correction

In ideal models there is no concept of errors. However, realistic self-configuring systems must deal with various sources of error as are shown in figure 37. It shows one more component that is different from figure 38: The extractor / selector component. The job of this component is to actually fetch and transmit the information in Φ over to the (re-)configurator Θ. The speed at which this is happening is $n$ bits/s (as already discussed). What is new is the idea of a duplex channel that is used for two-way error correction. This second channel can be used to request defect parts of configuration description which could not be reconstructed by the reconfigurator despite transmitted redundancies.

The top line in figure 37 is showing a collection of typical error types to be expected in self-configuring systems:

- Deterioration of storage pool for reference configurations
- Transport and selection errors during extracting or selecting patterns
- Transmission errors between extractor and reconfigurator
- Errors in processing the configuration by the reconfigurator (bad unpacking, bad decoding, bad decompressing, bad optimization for constraints)
- Deterioration of plant's switching components (e.g. transistor noise).

The self-configuring system will employ all necessary error correction techniques that are available, including redundant storage, corrective information, corrective protocols, state monitoring and refresh strategies.

I believe that the model suits several technical areas and research disciplines which are dealing with configuration correction:

1. The study of Case-Based Reasoning (CBR) and related approaches can be understood in terms of dealing with errors while retrieving reference configurations. CBR researchers speak of "repaired cases".

2. Another domain that also fits into this model is cybernetics. The closed-loop controller implements a basic two-way correction protocol that is counteracting strong deterioration forces in the plant.

3. In classic communications theory, a communication channel is a special case of the model in figure 37 and only carries error correcting information in one direction. Although it is not concerned with configurations, configurations can be understood as *signals* when transmitted sequentially. In communication technology the system plant is yet another storage system (RX buffer). This allows cascading of such systems.

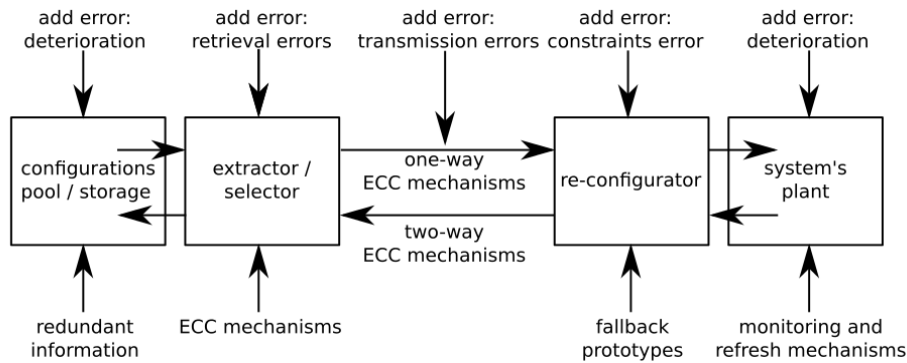

*Figure 37: Various error sources and error correction mechanisms (redundancy, generalization, monitoring & reprint).*

## 4.5 Summary of Chapter

In this chapter the purpose of the reconfiguration controller was discussed. The controller manages the selection and evaluation of potential plant modes in comparison to the plant mode already installed on the plant. The term *mode* has been used to describe different configurations with more distinct effects and which are changed only infrequently. However, a controller can choose to reconfigure the plant partially and continuously. In that case the controller update strategy becomes integral part of system's behavior as will be discussed in more depth in chapter 7 on p. 37.

The heart of a risk-minimizing controller is the mechanism for evaluation of configuration transition strategies (or reconfiguration strategies). Those strategies define when and which configurations will be deployed on the plant ("installation of configuration"). Main risk from such strategies are that they require more time than is available or more resources than are available.

The configuration selection process can mix several techniques in order to obtain a new configuration. Among them such techniques as stochastic, linear optimization, memory-based or mixture-based approaches. Some of these methods do not always require an explicit storage but their logical preference (bias) for producing certain type of configurations can be understood as biased access pattern in virtual storage (deformed space of configurations to select from).

The controller is also involved in correcting errors introduced in the process of retrieving, unpacking and implementing configurations. Controllers equipped with rich error correction mechanisms can exhibit corrective behaviors known from closed-loop control.

Not discussed in this chapter was the role of criteria for configuration selection. Configuration controllers could rely on parameters which would describe the "goal" or "regularization terms" for its optimization functions. However, the same can be achieved by reorganizing the storage: A controller would express the same optimization process as different strategy after storage has been reorganized. This approach is preferred here.

# 5 $\Omega$-units

## 5.1 Introduction to $\Omega$-units

In section 2.3 I have discussed that, given a range of popular configuration domains, representing configurations as a pattern in a configuration space is not implausible. I will further assume that this is generally theoretically possible even if not always easy to attain practically.

So far, I have mainly highlighted the static aspect of configuration representation in the sense of *configuration as state / structure*. Now, I would like to explore into the dynamical aspects of configuration, into the how?

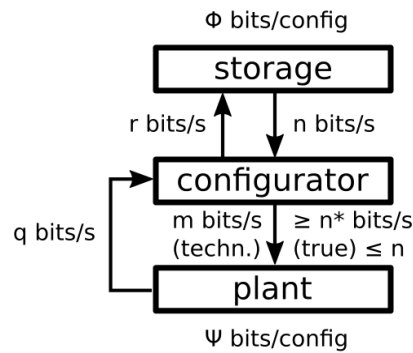

*Figure 38: The $\Omega$-unit: A general view on the reconfiguration process in a self-configuring system which is based on "link speeds" or "link bandwidths" in [bits/s].*

In figure 38 a general concept of a self-configuring system is displayed which I use to understand reconfiguration processes in systems. Its three basic components of storage, configurator/controller and plant comprise the standard $\Omega_{min}$-unit.

The heart of the model is the plant. It contains a configurable system as was described in chapter 3. It can be a single scalar value (e.g. a simple closed-loop control scenario) or a complex collection of matrices which represent connectivity of parts.

The configurable plant components must receive information in order to change their state. This model assumes that there is a limiting plant configuration bandwidth of $m$ bits/s. This value is a bulk rate. The net rate over this link can be much lower ($n$ bits/s).

The *n* bits/s come from the link speed between the *storage* and the *configurator*. The relationship between n and m is like between compressed bitstreams (*n*) and uncompressed bitstreams (*m*). The values *n* and *m* limit the speed at which a plant can perform a reconfiguration in real-time.

However, there are other limiting factors as well. For example, the value *q* [bits/s] limits the monitoring speed of the plant. Depending on how large the class of errors is, the time required to properly identify the right type of error and hence to suggest the right type of new configuration depends on this bandwidth. So, if there are 8 possible errors to be detected and *q* is 1 bit/s then it will take 3s in order to identify the problem with the plant.

Another limiting factor is the retrieval bandwidth *r* [bits/s]. This is the speed at which a configuration repository can be addressed. What does it mean? If the storage knows of 1024 configuration patterns then 10 bits of information must be provided in order to select one of them. If *r* is 5 bits/s then the selection time is 2s.

*r* and *q* limit the minimum time required to react to a problem. In the above example this would yield a 5 seconds delay (3s for monitoring and 2s for selection).

Let us assume that the required configurations $\Phi = \Psi$ [5] are of same size – for example 2048 bits and let us assume that $n = m$ is 128 bits/s then the reconfiguration duration would take 16s. In total, the time required to reconfigure the system is 19s (3s+2s+16s) and the system's plant is without reaction for 16s.

## 5.2 Various Kinds of Self-Configuration

Configuration of systems can be distributed among several components. Depending on where certain functions had been allocated we can discern several types of configurable systems (cf. figure 39).

When we speak of self-configuring systems, we normally imply self-3-capability. However, most technical systems belong to classes *a*, *b* or *c*.

---
5  These are the "sizes of patterns"

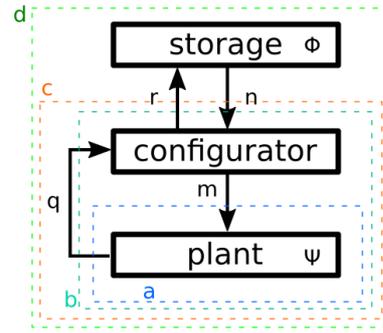

*Figure 39: Different types of configurable systems.*

  a) *Configurable*
  b) *Self-Implementing*
  c) *Self-Monitoring, Self-Implementing*
  d) *Self-3-Configurable*

The Ω-unit shall pose a minimum architecture for self-3-configurable systems. Since minimum models are often not directly observed in technical systems, extended Ω-units will be more common. In the following I will use the notation Ωxxxx in order to express how many layers are involved in the Ω-unit and how strongly each layer is fragmented. A Ω211-unit will have two storage sources, one controller and one plant. This notation is to be used with caustion because it assumes a certain organization, for example (Φ,Θ,Ψ,Ψ) but indeed could be also (Φ,Θ,Θ,Ψ). Whether such system constellations are convertible into each other is of high interest to engineers.

## 5.3 Summary of Chapter

This chapter was concerned with understanding self-configurable systems as units and as architecture layers of storage Φ, controllers Θ and plants Ψ. The standard Ω-unit is minimum complexity architecture for self-configurable system. However, nesting of such units among each other creates more complex Ω-units for which it must be decided if they introduce a genuinely new quality to the system or whether their differences are purely motivated by engineering mods – improvements designed to increase speed or effective storage capacity. In the next section I will elaborate some more on possible sources variability and analysis.

# 6 Engineering of Self-Configuring Systems

Figure 38 is showing a monolithic perspective on self-configuring systems. In that model the communication channels must support enough bandwidth in order to satisfy system's real-time requirements. Systems with easy requirements are found to be more like the monolithic model (one storage system, one engine for configuration, one plant). However, some applications must adapt quickly and the link speeds could be insufficient with a given technology. In that case we can see several basic strategies to reduce bandwidth demand through distribution techniques.

## 6.1 Fragmentation Exploits

The most obvious way to optimize to optimize reconfiguration speed is to reduce configuration pattern sizes $\Psi$ and $\Phi$ as is shown in figure 40. The replication of channels per plant is effectively doubling the transfer speeds by a factor of 2. The precondition for being able to split up a self-configuring system like that is the independence of plant fragments $\Psi$ from other fragments of $\Phi$. This would be naturally the case for all kinds of homogeneous array plants, like e.g. screens.

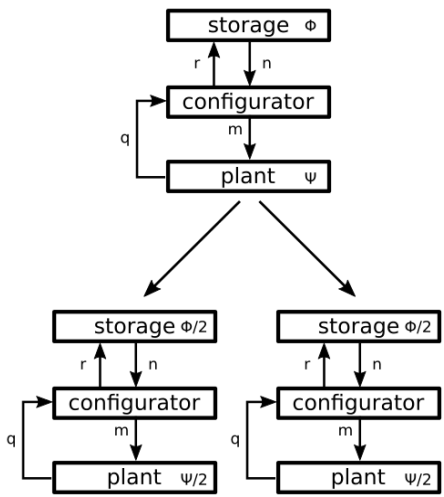

*Figure 40: By dividing the configuration space into two systems, the size of $\Phi$, $\Psi$ is reduced by half. This yields improved reconfiguration performance by factor of 2.*

## 6.2 Prioritization of Implementation Activities for Optimized Use of Bandwidth m

Another way to optimize for lacking bandwidth is to prioritize its use as is shown in figure 41. The plant is divided into several areas (need not be continuous) and ordered. The configurator uses the link channel to first service high priority areas before servicing low priority areas. This improves a system's graceful degradation characteristics. In case of increased reconfiguration demand the system will be performing best efforts to keep pace with the demand.

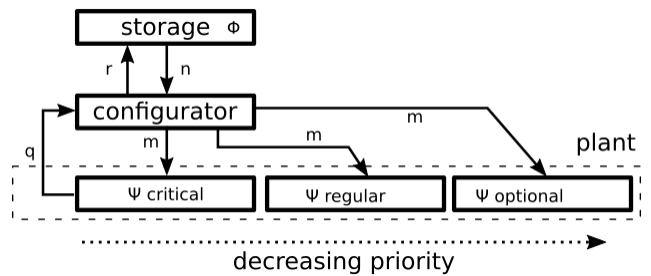

*Figure 41: Better allocation of bandwidth m in respect to system's mission.*

This approach does not require to have exactly three priorities and thee plant fragments. The plant fragmentation can be pursued into infinitesimal micro-scales. In that case there is up to an infinite number of priorities. Bit fragments arriving first are used to update plant fragments with highest priorities as shown in figure 42.

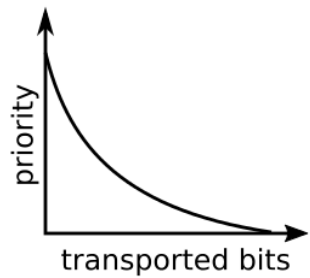

*Figure 42: A qualitative curve describing a realistic servicing characteristic for infinitesimally fragmented plant.*

## 6.3 Prioritization of Retrieval Activities for Optimized Use of Bandwidth n

A similar approach can be taken for optimizing the utilization of bandwidth $n$. However, the criterion for storage fragmentation is different from the plant: The amount of shared information (i.e. commonality of information) in configurations. The rationale behind the choice of such a criterion is the assumption that the environment and the system state evolve under some kind of inertia. In that case configurations used in temporal proximity should require small deltas.

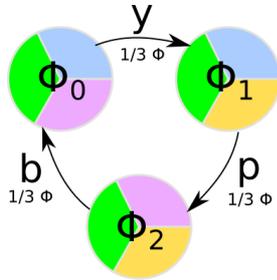

*Figure 43: Cake diagrams showing delta-upgrades of configuration. Green fragment is never transported, fragments y(ellow), p(urple) and b(lue) only when needed.*

Under this assumption it is smart to access only Φ fragments which are only the difference between the current and a new configuration (cf. figure 43). This can significantly reduce the effective number of bits required to reach a new configuration. This improvement is the smaller the greater the configuration change is to be performed because it is more likely that overwriting the more common configuration parts is inevitable (cf. figure 44).

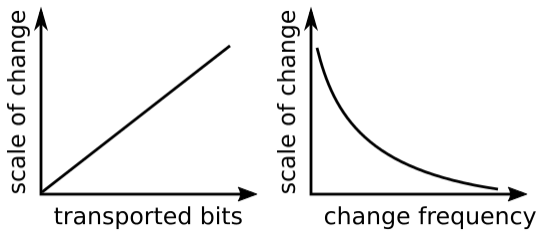

*Figure 44: Rationale behind segmentation of storage by demand frequency.*

An architectural model for this approach is shown in figure 45:

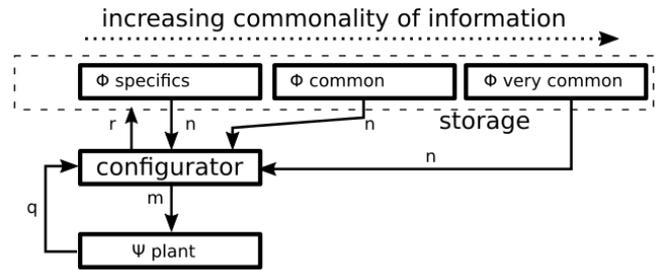

*Figure 45: Better allocation of bandwidth n in respect to system's state and environment's state.*

## 6.4 Compression of Configuration Storage as Means to Optimize Use of Bandwidth n

Since the configurator accesses the storage with a maximum bandwidth $n$, it seems desirable to minimize the size of Φ in order to improve reconfiguration performance. One way to do it is to compress the configuration pattern and to decompress it when needed.

There are two basic flavors of compression: loss-less and lossy compression. If the goal is to save as much space as possible then configurations should be compressed with lossy algorithms. How lossy (or loss-less) a compression may be will depend on the intended effect. For example, in MP3 a human auditory model is used to determine which pieces of information can be lost in certain parts of an audio track. Another parameter is for example the bitrate factor which is depending on the question if the audio stream is for a preview or for high quality streaming. Thus the main question is simply whether compression is good enough to the effect. In plants where small deviations will endanger the plant efficacy, lossy compression must be close to loss-less. Frankly, in such situations algorithms for loss-less compression should be preferred.

Even if lossy compression strategies can yield pretty good results, the question is whether we can reduce the size of compressed configurations even further. If we briefly think back of sections 4.4 "Error Correction" and 6.3, another idea to reduce the size of configuration patterns even further could be to rely on "repair mechanisms" of the system in order to remove repairable part of the configuration. The amount of update is normally defined as the difference between the reusable amount of configuration and the compressed size of the configuration. By dropping data which can be repaired, this delta can be reduced even further (cf. figure 46).

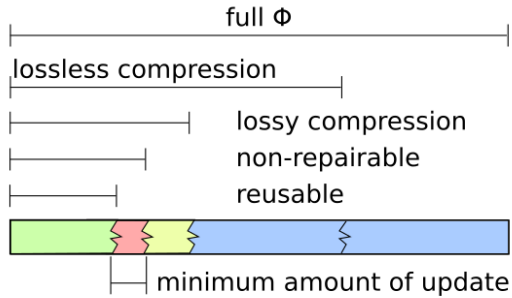

*Figure 46: Identifying the minimum necessary amount of transportable configuration*

The idea is shown in figure 47: The configurator accesses very sparse partial representations of Φ in order to assemble a compressed version of it ($\Phi^C$). This yields a configuration with a recoverable error (lost part from lossy compression) into the configuration that must be corrected through interaction between adaptive algorithms and postponed systems' (Σ) feedback.

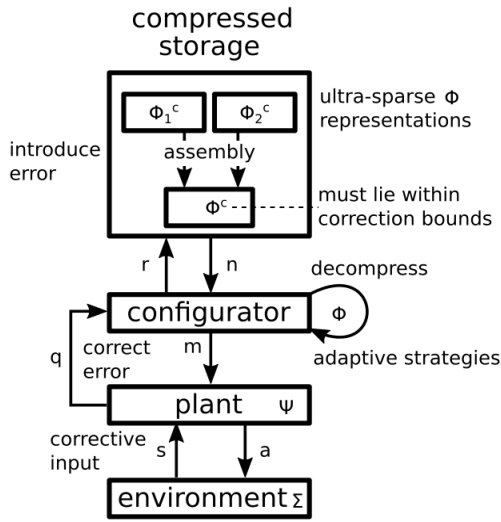

*Figure 47: If the configurator can access and assemble configurations from partial, ultra-sparse representations and then decompress them, then effects of bandwidth n on system's reconfiguration performance can be significantly reduced at the expense that the system has to perform corrective adaptations of the plant configuration.*

Access to sparse representations via r-selector is remarkably similar to accessing abstract concepts: A minimalistic configuration fragment is an information atom which role is solely defined by its address (r) and its assembly rule structure (which is more commonly known as an *ontology*).

Since we see configuration assembly now for the second time (at first we saw it in figure 45) the suggestion is near that storage fragmentation should occur not only horizontally but also vertically, more specifically at least hierarchically in order to:

a) minimize effects of limited bandwidth n

b) maximize effective storage capacity for configuration patterns

A resulting system design would be characterized by the following properties:

a) The system would maintain a hierarchical storage of patterns for configurations

b) The system would update its working configuration quickly in a prioritized, possibly opportunistic manner.

c) The system would expose a mixture of behaviors ranging from knowledge-based open-loop to adaptive closed-loop control.

d) The system would expose features of graceful degradation.

## 6.5 Clustering of Configurations for Optimization of Use of Bandwidth r

So far, no considerations were made for optimizing the influence of bandwidth *r* on the reconfiguration performance of a self-configurable system. In fact in a plain configuration table (LUT – look up table) the requirements toward bandwidth are really small. With a selector of 128 bits of width, it is claimed to be possible to select all atoms in the universe.

This number of bits is relatively small but the true bandwidth requirements rely on a few more things:

1) How often must the system reconfigure?

2) Does the system perform configuration assembly?

Both factors can greatly increase the required bandwidth for the *r* link.

If the selection bandwidth *r* ever becomes a limiting factor then this will pose an obstacle for adding more reference patterns to storage because the address sizes raise. In order to keep accumulation of further configurations feasible, one way to optimize bandwidth *r* is to switch from absolute to relative addresses which are (hopefully much) shorter. In order to really benefit from such redesign of address mechanisms, it is necessary to organize related configuration patterns in close proximity in order to minimize the number of bits required to describe the new address for access.

The allocation requirements for patterns (which should not be placed randomly) will lead to contextual clustering of configuration patterns. Some areas in the pattern memory will look like a collection of very similar patterns because they are frequently used in temporal proximity by the reconfigurator. This should be especially true for all low level sensorimotor configurations. This type of cluster is excellent for performance optimization as was proposed in figure 45: In such clusters the amount of information necessary to transfer a configuration can be greatly reduced if common information is not transferred.

However, there will be also clusters of very different looking configurations for the same reason: If the reconfigurator has to frequently switch between very unrelated configurations then it will like to have them closer in the storage system despite their dissimilarity. This should be particularly true for more abstract, partial configuration patterns used for assembly of configurations. Fragments used more frequently for assembly in configurations need to be located more closely together. This in turn would propose that the equivalent of "concepts" (which are the partial configuration patterns) need to form something of a semantic cluster. This could explain why systems could auto-associate certain concepts faster when presented with certain other concepts before (priming [28]).

## 6.6 *Optimization of Bandwidth q*

The remaining communication channel to consider for optimization is channel *q*. This channel is used for monitoring the system's plant performance.

The system plant is permanently interacting with postponed systems which is normally the environment Σ. There is a bidirectional interaction between the plant and Σ conveying *s* bits of feedback to the plant and *a* bits of feedback to the environment. If the plant is synchronized with the environment then no information is conveyed over these channels (cf. figure 48). Please note that channels *s* and *a* are not synonymous with sensors and actors, despite that most technical implementations look this way. The transfer rates obtained on sensors and actors are only theoretical upper bounds. Practical upper bounds are typically far less.

The goal of monitoring the plant Ψ is to detect if the system plant is not capable of keeping bitrates of *s* and *a* close to zero.

1) High *a* means that configuration of plant Ψ is not in sync with Σ: Expends energy in order to guide Σ.

2) High *s* means that plant effect is not achieved.

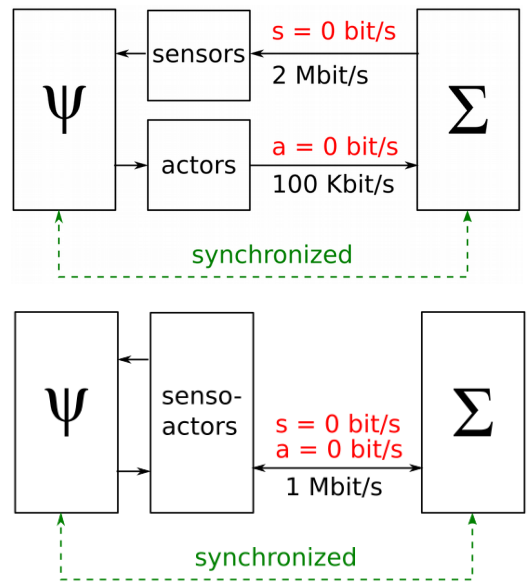

*Figure 48: Bitrates obtained on channels s and a do not necessarily represent the technical capability to transfer bits.*

Information conveyed by *s* and *a* is not directly available to the reconfigurator. A derived piece of information *q* (in sense of "quality") is transferred to it in order to choose a better configuration.

I see two main ways to understand *q*: a naturalistic way and a subsidiary model suitable for computers (cf. figures 49 and 50).

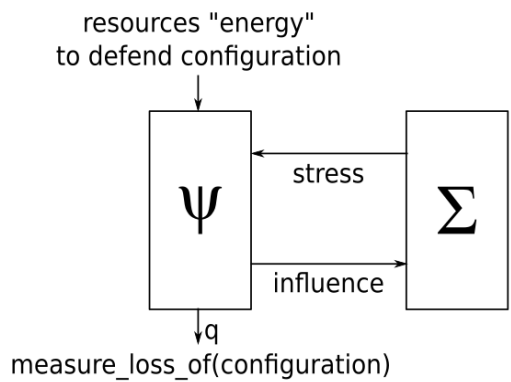

*Figure 49: Naturalistic model to measure q*

Figure 49 assumes a physical interaction between Ψ and Σ. Σ will induce disorder in Ψ depending how violently Σ evolves and how many resources are used by the overall system in order to defend the configuration in Ψ. Measuring the amount of loss of configuration in bits (divergence) yields a total measure for *q*. In analogy, any fragment of configuration can be measured for *q*.

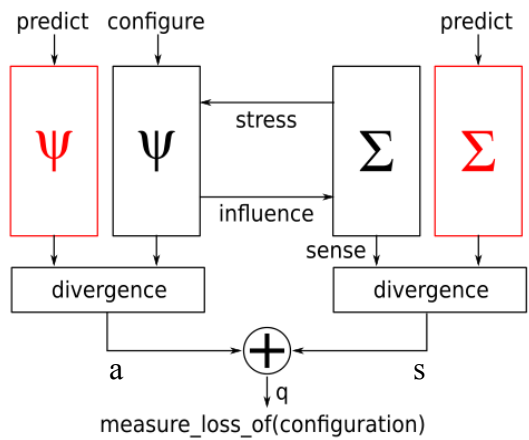

*Figure 50: Subsidiary model to measure q for computers.*

Since computer programs do not get "disordered" by external failures, due to the way digital hardware is built, a model based on comparing predictions with true evolution of systems must be depended on in order to measure the performance of the plant. Figure 50 shows how divergence (e.g. Kullback-Leibler divergence) could be obtained for plant and environment in order to compute a *gross-q*. Examples of such approaches are found in [29], [30].

The above models suggest a gross-*q* value for measurement but indeed a scalar *q*-value offers little information about *how* to reconfigure, only *that* to reconfigure. A scalar measurement of *q* can yield important argument to allow larger reconfigurations and to accept longer reconfiguration delays. However, this would unlikely pose a serious bandwidth problem to most systems.

Things will look differently if the monitoring performed shall also contribute to selecting a new configuration much better than randomly. In that case monitoring must cover performance evaluations at a more granular level in order to understand how the plant fails and how this relates to configurations.

Depending on granularity of plant surveillance, the amount of information collected on the q-channel can now significantly grow. Moreover, if the collection information is a vector or large matrix of *q*-values (*q*-maps) then the system must map between detected position of divergence[6] and new addresses used by configurator for *r* (retrieval). This sounds very much like a cognitive task but indeed is just an arbitrary mapping (function) for this model. It could be a computed function or trained function but is a function nonetheless.

If we assume that the mapping between *q*-maps and address space can be prioritized, that not all obtained *q*-values are relevant to the current context, then those could be avoided to be collected. If the mappings were part of the actual configuration then we would get a "contextual attention" for the system (a cognitive interpretation). This again would achieve a reduction of demand for bandwidth on the *q* line. Idea is visualized in fig. 51.

Indeed, this concept is very near to common computing technology: An interrupt mask is set by program procedures. Remaining allowed interrupts can trigger a jump to new procedures referenced in an interrupt vector

---

6   It is the "problem position", so to say

table which can set a different interrupt mask. However, in computers this is not mainly used to keep interrupt rates low but an interplay between a Kernel scheduler and interrupt system exemplifies how CPU performance can be channeled towards certain tasks.

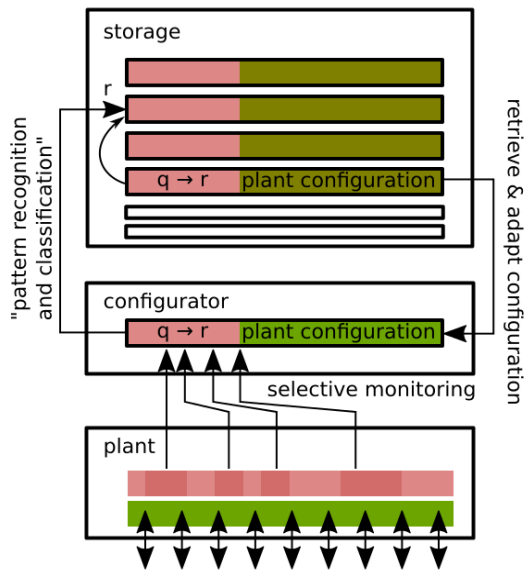

*Figure 51: Selective plant monitoring based on a qr-map provided along with the plant configuration. A similar approach is taken in CPU architectures where a µ-Code is defining the actual operation in the CPU-plant.*

## 6.7 Refactoring of Configurations and Systems

### 6.7.1 Clarifying the Meaning

What does it mean to *refactor* configurations? Since this term is intuitive for the moment, maybe it is a good idea to start with a well known example to get an idea what this could mean. Let's take a polynomial function.

Polynomial functions can be transformed into a Horner schema. By this transformation, the number of multiplications necessary to compute the polynomial is greatly reduced.

Removing a polynomial element from the function is also very easy through polynomial division. By pulling out polynomials the resulting product branches *permeate* the other parts of the statement. The benefits of this are simple decisions regarding certain curve properties (poles, zero crossings, etc.).

The question is then if it is possible to do such refactoring also on configurations and related behaviors with similar benefits?

What could we expect from refactoring? Normally, refactoring brings non-functional improvements such as improved speed, energy efficiency, reduced maintenance cost or a higher reuse rate.

How can these ideas be applied to configuring systems? Firstly, I see two parts to it: The static configuration and the resulting plant behavior. At minimum, the plant behavior is static again. Refactoring of configurations is mainly about pulling behaviors (entangled sets of final static configurations) out.

### 6.7.2 Decomposition as First Step of Refactoring

Consider topmost behavior in figure 52: It is a single plant configuration in input (configuration) space and the plant produces a sequence of configurations in the output (configuration) space. Given the mission of the job, the produced behavior (in this case a cyclic attractor) is an approximation of an ideal behavior (fine line). If the system wanted to approximate ideal behavior but is stuck with a non-decomposable plant then the approximation can be very limited. Imagine that the behavior can only be configured in terms of width, height and traversing speed.

Now, the goal of refactoring is to gain additional configuration opportunities. This could be done by finding piecewise functional substitutes (possibly approximations) of the original monolithic plant behavior as is seen in the middle of figure 52. This could be achieved by splitting the plant in four sections (and rationalize bandwidth $n$, as was proposed in figure 40) or to create some kind of plant overrides. The actual ability to do the second will greatly depend on the plant's design. In any case, the result would be that now 12 (4x3) configuration parameters exist which can be used in order to optimize the plant performance in any of the four behavior sections.

Decomposition can continue by creating even smaller fragments of behavior. In figure 52 this is shown at the bottom. The refinement need not to occur "flat", i.e. the meta-modes remain in place and remain accessible for "fast configuration" cases. The children configurations can be expressed relatively to the super configurations which yields advantages in term of adaptation performance and semantic contextualization of the parts.

### 6.7.3 Optimization of Parts as Second Part of Refactoring

However, one issue must be considered: Proposed behavior fragmentation is introducing a mode transition problem. Stretching one of the partial behaviors leads to a "broken line" - a discontinuous development of configurations. Plants designed with very narrow policies, like for example shown in figure 23 on p. 15., will stop behavior at a broken boundary if program is not contiguous.

For plants producing continuous behaviors this can be obviously a problem. In order to overcome such mismatch-continuation problems each factored out behavior must not only be simpler but also more resilient. As we have discussed on p. 16, adding corrective fields is making the plants behave more resiliently. With each step of refinement, i.e. by continuously simplifying the fragment behaviors and by making them more resilient, we should observe a generalization of policies towards uniform field-like sections. If some kind of reuse mechanism is implemented on the system which would prevent approximation of the worst-case plant, as was described in chapter 3, the system would start to assemble basic, universally applicable operators. At the end of this process the system should explain itself as a set of piecewise linear models or other "trivial" models which we could call *theories* if the system used them to simulate its performance ahead of time.

### 6.8 Compilation of Dedicated Plants as Last Part of the Refactoring Process

Last but not least, there is the risk that activities of the reconfigurator associated in organizing the necessary mode switches is still not performing enough. In that case the reverse operation would complete the refactoring process: The compile.

In figure 53 the complete process is shown: So far discussed, I have motivated to dissolve the plant behaviors into more basic behavior,

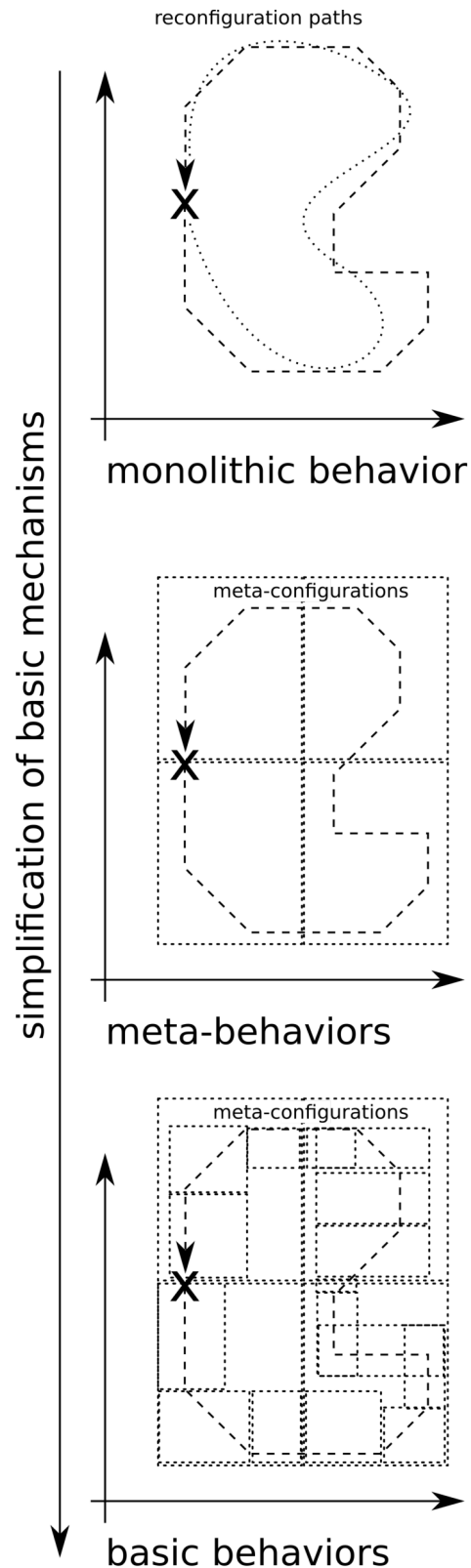

*Figure 52: A system can be decomposed into various more basic configurations by introducing meta-configurations.*

each with its own set of configuration parameters. These piecewise configurations can imitate the original plant by reconfiguring from one behavior into another one. This offers additional opportunities to optimize the behaviors. However, because the re-configurations could be too slow for the system's final environment, the system must *compile* a new integrated plant which has a different set[7] of configuration parameters than the original one. It performs an optimized behavior at higher speed but reduced flexibility when compared to the piecewise plant approximations.

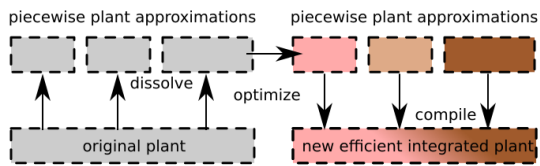

*Figure 53: Refactoring of behavior by dissolving, optimizing and re-compiling the plant*

More semi-formally, a unit made of $\Phi+\Theta$ impose a transition field (a "vector field") on the $\Psi$-plant's potential configuration space. $\Phi+\Theta$ can be replaced with any more complex cascade of ($\Phi+\Theta$) which at minimum is capable to reproduce (or approximate) initial configuration transition sequences (behavior). After that transformation behavior can be optimized. If resulting new behavior can be reintegrated, some of the $\Phi$s and $\Theta$s get removed for sake of better performance or cost.

### 6.9 Operations on Ω-units

Technical systems consist of many components, each showing capabilities in ranges as shown in figure 39. A typical question when dealing with distributed systems is whether they can be transformed into other constellations of systems (example question in figure 54).

There are of course two questions to be answered first:

1. What is the language in which the architectures can be expressed?

2. What are the rules for transforming and comparing the expressions?

Currently, there is no such language or precisely defined rules for manipulation, however the idea is strongly motivated by the block algebra for control systems[8].

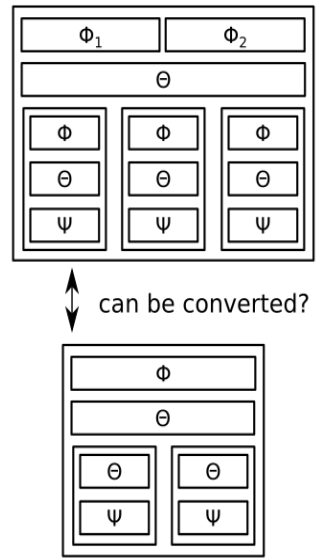

*Figure 54: An example problem: There are two system architectures for self-configuring systems. Can one system be refactored into the other system without losing any significant properties?*

Figure 55 is showing a line of transformations where a single, integrated Ω-unit is decomposed into two smaller integrated Ω-units. This decomposition is played through in order to get a first idea of what operations could be performed on the graphical representations based on nested components. The representations do not contain all possible com-munication links *q*, *r*, *n*, *m* but assumes that they exist in a combinatorial fashion between layers. This simplification could be miss out on some important properties but since this is a first idea for further exploration the communication links have been left out.

---

7   Often a reduced set of coefficients

8   http://www.msubbu.in/sp/ctrl/BD-Rules.htm

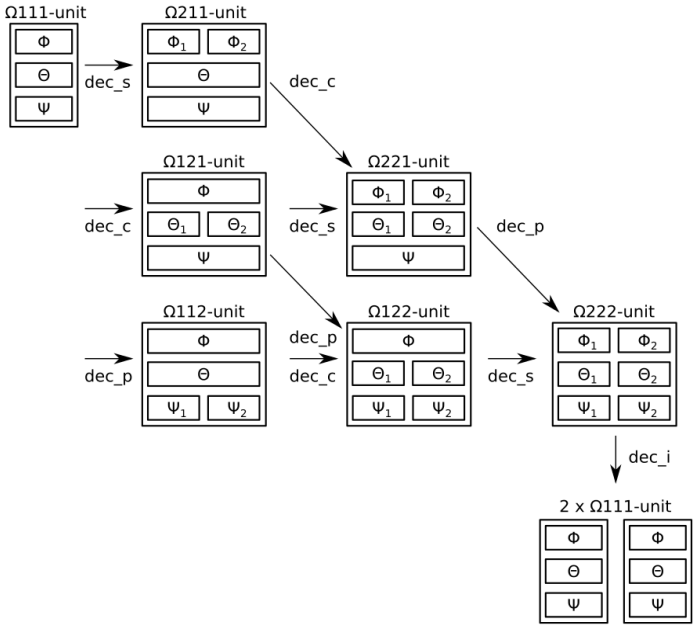

*Figure 55: Decomposition paths by identifying independent areas of storage, control or plant.*

There are four operations shown in the decomposition diagram of figure 55: *dec_s*, *dec_c*, *dec_p* and *dec_i* which are explained as follows:

**dec_s**: Decomposition of storage. This is equivalent to segmentation of the input configuration space. A separation could occur between L- and N-space but such separation would not be used in practical engineering. More practical separations would be between main modes: Storage encoding would be adapted to shape of legal space relevant for a particular mode. Modal separations could be accompanied by dropping of configuration space dimensions (along which the policy is constant). It is also possible to think of separation of aspects. This could be practical in situations when the storage is externally updated per aspect via different paths of configuration space remodeling. Operation dec_s is permissible if decomposition does not lead to losses of storage content, if redundancies are introduced or if losses can be corrected by error correction mechanisms.

**dec_c**: Decompose controller. A controller could be decomposed, for example because certain dimensions of plant can be optimized independently. This would be the case, for example for a controller based on linear calculus where each dimension depends on all inputs from storage but output dimensions are calculated individually. Decomposition of controllers is typically introduced for performance reasons (parallelization). Operation dec_c is only permissible if the output of controllers does not require a postponed integration in order to be deployed on plant or if the errors are so small that corrective plant behavior (corrective fields) will overcompensate introduced errors. Introducing redundant instances of $\Theta$ is not a decomposition in sense of dec_c but a superscript index $\Theta^x$ can help to understand involved failure characteristics.

**dec_p**: Decomposition of plant. There are many plants which pose an integrated unit and hence must be provided with all necessary arguments in order to execute their policy. However, many other plants have array characteristics or perform the same action on sequences. In that cases parallelization of plant is means for increasing plant performance. Such parallelization is accompanied by replication of *m*-channels. The controller sends a single copy of $\Psi$ *m* which arrives at all plants replicated (broadcast). If controllers had been decomposed, too, then each of the plant replicas receives information from both.

**dec_i**: Decompose into new units. Once storage, control and plant have been sufficiently decomposed, opportunity arises to split up the system in completely disjoint $\Omega$-units for replacement. This rather trivial operation can not be permitted if any plant fragment $\Psi_x$ has transient dependencies to storage fragments $\Phi_{y \neq x}$.

As has been remarked several times, self-configuring systems can be assembled from smaller systems. If a plant is a self-configurable unit on their own then we are dealing with a nested $\Omega$-unit (left pictogram in figure 56).

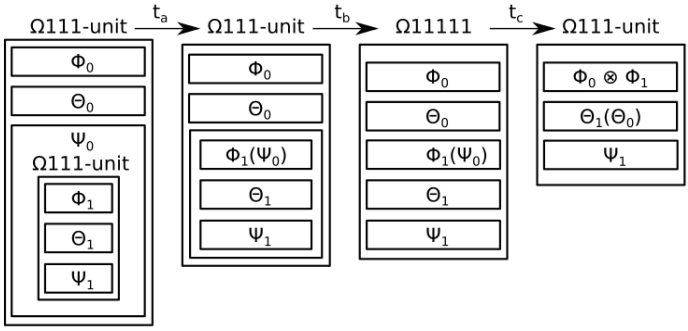

*Figure 56: Integration of nested self-configuring systems $\Omega_1(\Omega_0)$ into a new integrated unit $\Omega_3$*

The three transformations $t_a$, $t_b$ and $t_c$ (shown in the upper part of figure 56) represent three integration phases. Currently, those transitions do not suffice precision requirements to make them formal operators but following them is a common engineering practice.

The starting scenario is a single nested $\Omega_1$-unit in another $\Omega_0$-unit. The nesting occurred in the plant $\Psi_0$ of the outer $\Omega_0$-unit.

*Connecting Transition $t_a$*: In this step the two systems are "hooked up" in order to work together. The configuration storage $\Phi_1$ is made dependent on the plant configuration $\Psi_0$. In ideal cases $\Phi_1 = \Psi_0$ as would be true for *pure communication nesting*.

*Dissolving Transition $t_b$*: The functional components of $\Omega_1$-unit get unframed – they wander into the realm of a new $\Omega_3$-unit with more sophisticated structure. Remarkably, it yields a five-layer unit ($\Omega 11111$-unit) with two units of storage and two controllers. The controller $\Theta_1$ and plant $\Psi_1$ have become essential parts of $\Omega_3$.

*Consolidation Transformation $t_c$*: Since, in theory, the controller $\Theta_0$ does not alter configurations on the way between $\Phi_0$ and $\Psi_0$ it seems to be reasonable to move the effective middle-way storage $\Phi_1(\Psi_0)$ to the top where it becomes integrated with the original storage $\Phi_0$ in some form of an outer join operator $\otimes$. This operation will normally be executed as adding dimensions to storage and combinatorial configuration expansion. In practical cases, such space is eligible for strong compression and/or refactoring for independent input configuration clusters.

The construction rules are not defined but certain basic formalisms could be used:

$\Phi = \Phi, \Psi = \Psi, \Theta = \Theta$ (are of same power)

$\Phi + \Theta \neq \Theta + \Phi$

$\Omega_{min} = \Phi + \Theta + \Psi = \Phi + (\Theta + \Psi) =$
$\qquad\quad (\Phi + \Theta) + \Psi$

$2\Omega = (\Phi+\Theta+\Psi), (\Phi+\Theta+\Psi)$

$\Omega^2 = (\Phi+\Theta+\Psi)(\Phi+\Theta+\Psi)$
$\quad\; = (\Phi+\Theta+(\Phi(\Psi)+\Theta+\Psi))$

$(\Phi+\Theta+\Psi) = (\Phi+\Theta+\Theta+...+\Theta+\Psi)$

$\Phi+\Psi$ : forbidden operation

*(some sketchy propositions for further elaboration)*

### 6.10 Summary of Chapter

This chapter was concerned with engineering of self-configurable systems at an architectural level. The role of link speed $q$, $r$, $n$ and $m$ has been discussed and how it could be improved. Optimization of system architecture can yield many similar variants and all of them could be not obviously related to the minimal $\Omega_{min}$-unit.

Since architectural design aspects play an interesting role in comparing systems among each other, I let myself inspire by the block diagram transformation rules in control theory in order to come up with a sketch what kind of $\Omega$-transformations we could expect and what way they could be formalized. I would assume that a fully developed $\Omega$-algebra[9] for self-configuring systems would build on discipline of Mereology [31].

So far, applying transformations is relying on deep knowledge of system at hand and individual talent to apply them. Important measure of quality of a finalized formal $\Omega$-framework will be the demonstration of capability to integrate other mathematical formalisms describing specific systems under design. Only then such a formal concept is going to become a reliable tool for analyzing and assessing system architecture decisions.

---

9   Not to confuse with omega algebra (established mathematical concept) which is an extension of the Kleene algebra.

# 7 Reconfiguration and Behavior

## 7.1 Iterative Reconfiguration and Behavior

Since, in theory, a system can reconfigure not only at major occasional events but can also perform more frequent configuration updates, reconfiguration becomes an integral part of a system's behavior.

In a mechanical system, each static layout of the components can be understood as a configuration. A dynamic evolution of this system is then understood as process of permanent reconfiguration. This means that the basic units of any dynamically evolving (self-)configurable system are its achievable static configurations.

Figure 57 depicts this idea and also shows bi-modal behavior and possible transitions between two behavioral modes (mode 1 & 2).

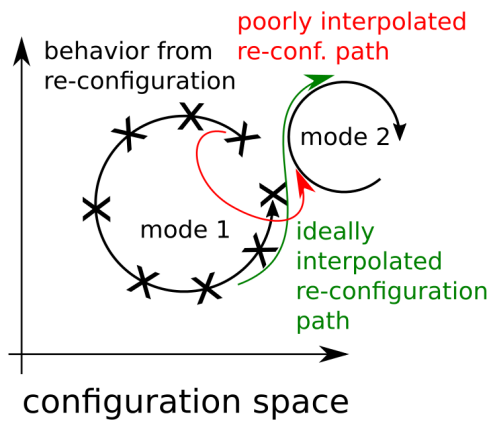

*Figure 57: Behavior as history of reconfigurations. A system can have meta-configurations producing characteristic reconfiguration patterns (behavioral modes).*

This example also shows how scheduling of mode switches can influence plant performance: The red line shows an immediate (greedy) mode switch. This switch forces the system to deviate strongly from established configuration flow – a situation associated with higher energy expenses on robotic systems. The green line represent an ideal transition path between the modes where the mode switch is scheduled "just right".

However, the question is then, how static configurations relate to plants where the configuration is controlling large sets of static configuration classes (=complete behaviors).

This process could be explained in terms of unsuccessful reconfigurations or excitations (excitations a,b in figure 58). As long as $q$ is reporting any kind of "problem" with plant performance, the result can be that a new optimized configuration is computed. This can cause a sequence of reconfigurations which would be perceived as self-motivated behavior from the outside. In figure 58 this is the self-propelling loop.

It is generally understood that self-excited behavior is associated with systems with internal feedback moving downwards their virtual energy landscape. At the end of this process most systems terminate in a trivial "energy sink" - a point attractor – and in few cases they are caught in less trivial cyclic behaviors – cyclic canyon attractors. This is also true for configurator-plant-models.

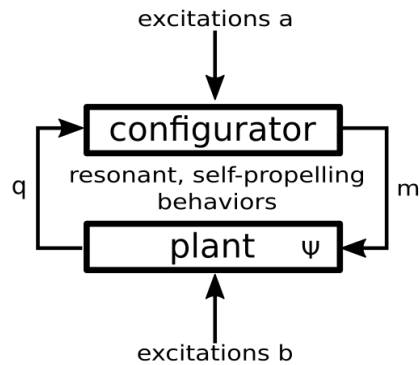

*Figure 58: Behavior as self-propelling process of reconfiguration.*

Continued activation is often caused by recurring excitations of type a and b:

a) changes to storage content or organization

b) changes to environment which take effect on the plant

When combining excitations with an iteratively optimizing configurator, the observed behavior could be potentially an infinite sequence of behavior without a clear final state. Since the resonant behavior of plant is repeatedly altered, continued excitations could bring the system into a state of overdriving its output by accident. We could call it an "erratic" state. The system should therefore posses the ability to detect and dampen its reconfiguration activities and bring down the plant in a recovering configuration.

## 7.2 Dynamical Constraints and Varying Behaviors at Different Reconfiguration Velocities

Aside of delaying mode switches in order to reduce the amount of performed reconfiguration (such as seen in figure 57), there are other dynamical aspects influencing reconfiguration paths such as dynamical constraints.

A very common problem in self-configuring systems is the problem of minimum configuration speed. Certain reconfigurations can be done at a high velocity but cannot be done below a certain threshold. When system's reconfiguration is used to interact with the environment and to exploit the resulting behavior for achieving a change in the environment, this is particularly true: Imagine swinging! Swinging the feet to slow or too fast will not maintain a swing-state in the environment!

But let us concentrate on the plant alone: Assume a situation in which three components must be *all on* or *all off* in order to guarantee a stable plant state. Let us further imagine that the system will collapse if the components are not *all on* or *all off* for more than a second. In that case the system can iteratively change component states from *all on* to *all off* as long as it finishes its transition within 1s of duration. This means that a system experiences a configuration space fragmentation that is depending on its reconfiguration dynamics. If the reconfiguration speed is not sufficient to flip all three states in a second then the system cannot transit between the modes..

This situation can be generally understood in a game-theoretic way: Any component has a certain opportunity to advance its state but if it fails to exploit the move then a "systemic adversary" will use the time (next moves) to advance its strategy. If the players (the components) cannot synchronize their activities in order to defeat the "adversary" then they can get locked in a Nash-equilibrium: They cannot improve their state further (toward the target configuration) despite that it is theoretically possible to be in the final configuration (optimal state). The exact boundaries of the Nash-excluded areas depend on the reconfiguration speed of the system and some characteristics of plant and environment.

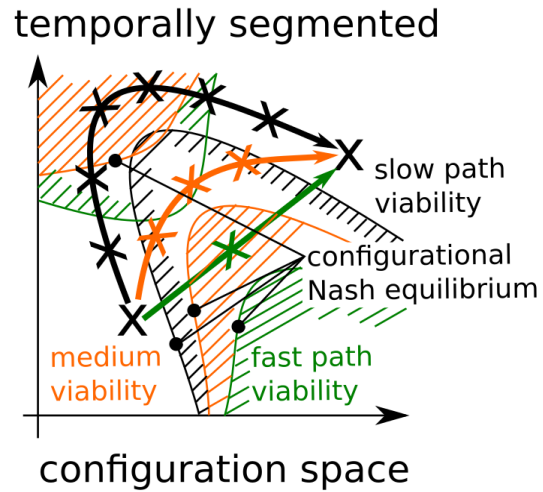

*Figure 59: Slower can be a lot slower!*

Figure 59 visualizes the problem: The green line is the fastest path of reconfiguration between the initial configuration and the final configuration. The green system is capable to change a large portion of the configuration within a unit of time. The orange system is a little bit slower. Selecting configurations within the area of the green intermediate configuration would harm the orange system – it must choose a circumvention for that area. The black system is even a little bit slower than orange and can only attain final configuration only after taking long detours.

This has interesting implications for reconfigurable systems design: A system which is a little bit slower in technical performance can be significantly slower in behavior performance. It is therefore very important to exploit all technical techniques to make systems reconfigure as quickly as possible.

Ultimately, systems can by no means be fast enough to arrive at a new state by transiting. In that case the system must replicate itself in the target configuration. This is like space traveling: Since no human can travel the time necessary to arrive at a different star, it could be easier to replicate humans on the spot.

## 7.3 Reasons for Hierarchical Modes

With all said so far, it is a little bit easier to estimate the role of modal style reconfiguration in self-configuring systems.

1) Fast, superordinate reconfiguration managers interfere minimally with plant dynamics as they reconfigure the plant (spontaneously). Hence, the reconfiguration process is not considered an integral part of system's behavior (cf. fig. 60, bottom row).

2) Optimization of internal bandwidths by splitting complex configurations into partial configurations. The final configuration is assembled by adding as little delta-configuration as possible (cf. fig. 47).

However, I see more reasons to follow multi-modal ("jumpy") reconfigurations: risk control!

Let us consider figure 61 for a while. The system behavior is defined in a hazardous configuration space. It cannot operate clearly off any dangerous conditions. Functions performed in red, blue and green area must be executed with high reliability. The simpler the behavior the more reliable it generally is. This can be explained in the amount of information that a configurator must add to the plant in ratio to the number of bits of information in the plant which can fall victim of entropic forces. The reliability R is better the higher the *m*. The reliability R is the worse the larger the ψ. The ε represents a technological unreliability factor.

$$R = \frac{m}{\epsilon \cdot \Psi}$$

$$[R] = \left[ \frac{bits}{s} \cdot \frac{1}{\frac{1}{s} \cdot bits} \right] = [1]$$

*Equation 8: Reliability of a system is a unit-less factor*

Since perfect reliability is difficult to attain, the question is how much is there "room for an error"? In this context, high safety means a sufficient *hazard key* as buffer between regular system configurations and illegal space segments in the amount of bits.

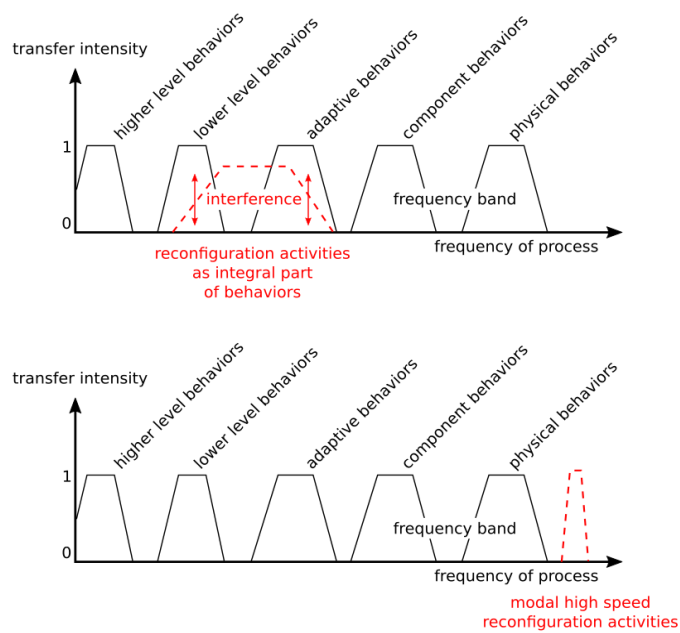

*Figure 60: Conceptual Spectrum: Reconfiguration mechanisms must be designed either as integrated or excluded part of behaviors. Above: Reconfiguration processes interact with plant behavior (e.g. cybernetics). Below: Reconfiguration process is made to be as spontaneous as possible in order not to interact (e.g. networks).*

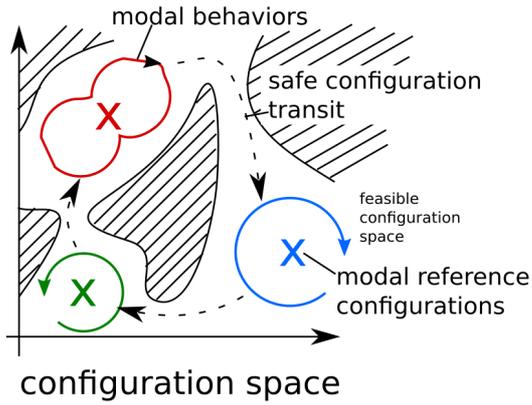

*Figure 61: A multi-modal reconfiguration example with safe transition behaviors.*

The more bits have to be added to the system configuration in order to reach illegal configurations, the more time is left for the reconfigurator to fix the behavior defects. Even if the idea of such hazard key lives of the imagery of a continuous behavior as is shown in figure 61, this is not necessarily required. All is needed that the selectable subspace for configurations (belonging to a modal behavior) has a minimum distance to hazard defined in bits. More formally, *L* is space of all attainable behavior-related configurations. *N* is space of all illegal configurations. The bits$_{diff}$ function computes a matrix of distances between all space points in *L* and *N* expressed in bits required for traveling from a specific *L-configuration* to a specific *N-configuration*. The minimum value $h_k$ defines the effective hazard key for that system behavior and is a measure of safety for meta-configurations responsible for creating the behavior.

$$h_k = min(bits_{diff}(L, N))$$

*Equation 9: Safety of a system behavior expressed as a hazard key $h_k$*

This measure delivers the "smallest fuse" even if a system gets extended with many additional "safe" behaviors. In that case, the safety of the system[10] is the minimum of all $h_k$ per individual behavior. The nature of this measure is that it remains stable even if the behaviors become reorganized into a new family of meta-configurations.

Please note that here the term *safety* means the safety from reaching illegal configurations. This is not synonymous with *product safety*. However, this concept is expandable to product safety if the configuration spaces *L* and *N* include state variables of the environment and if for this hybrid configuration a general cost function can be defined. In that case a hazard key $h_{kp}$ is measuring safety of a product in absolute terms in bits. But for a reasonable informational safety concept (*S*) it is advised to understand it relatively to system's reliability by multiplying the hazard key with system reliability R:

$$S = h_{kp} \cdot R$$

*Equation 10: Product Safety S [bits]*

The reason to treat the problem of safe distance for a self-configuring system is the question how the system can transit very narrow configuration bridges in order to attain new safe subspaces (as seen in figure 16)?

Well, the system could implement special transiting behaviors (or configurations of the plant which produce them) which are simple enough to be highly reliable. This would allow the system to iteratively update its configurations even through narrow configuration bridges. This implicitly forbids creation of monolithic behaviors with many free parameters which add to the system's risk of hitting *N*. Therefore, once the designer observes reaching invalid configurations by the system, his quite rational strategy would be to factor out fragments of behaviors out of the main policy and to approximate initial policy by introducing modal switching to the system.

## 7.4 Adaptation and Control

Systems presented so far have put a strong focus on the self-capability to reconfigure. This introverted activity seems to have a passive property: The environment changes and the plant has to reconfigure. However, reconfiguration activities of Ω-units are not necessarily only self-directed. Perfect self-direction is rather an interesting extremum in a spectrum of reconfiguration allocation.

What does the *reconfiguration allocation spectrum* mean? Let us think of an autonomous systems: In a "satisfied" state the resource inflow and outflow is at balance and the system needs not to alter its strategy. If the

---
10  representing all configurations obtainable by it system is layered then a satisfied state is characterized by the property that higher-level layers need not to be reconfigure while the lower-level layers could get adapted quickly. In ideal state the system needs not to reconfigure any of the layers. This is what is shown in the first row of figure 62.

However, at some point in time the system's environment or internal resources change so much that the inflow-outflow-balance becomes in-sustainable (second row in figure 62). As a result a greater reconfiguration is required.

Now, this reconfiguration can consist of any internal and external adaptations in order to restore resource balance (third row in figure 62). This can be at one extreme a perfectly internal process, like making a decision or a perfectly external change like pushing away chairs which are standing in the way. Pushing away the chairs would not change any internal policies. It would only adapt certain periphery policies responsible for performing the pushing.

How can this be explained in context of the here proposed view of reconfiguring systems? The question is justified because only the system's real Ψ plant is reconfigurable by the controller.

Well, whether the reconfiguration occurs mainly inside (e.g. because of strong disturbances) or outside is very much depending on the actual values $f_i()$ produced by Ψ. There is a *meta*-quantity missing for $f_i()$ in order to express how strongly a remote / postponed system Σ is subordinated to Ψ. We could call it *stiffness*. In theory, $f_i()$ can generate simply different values if more external influence is desired and the amount of influence is the properly selected by choosing the right output configuration *i*. However, at this moment it is not clear if this would be sufficient because stiff systems are also characterized by higher frequency control (or simulation) and according to the here drawn models this requires a change in coverage of q-monitoring and plant clock rates.

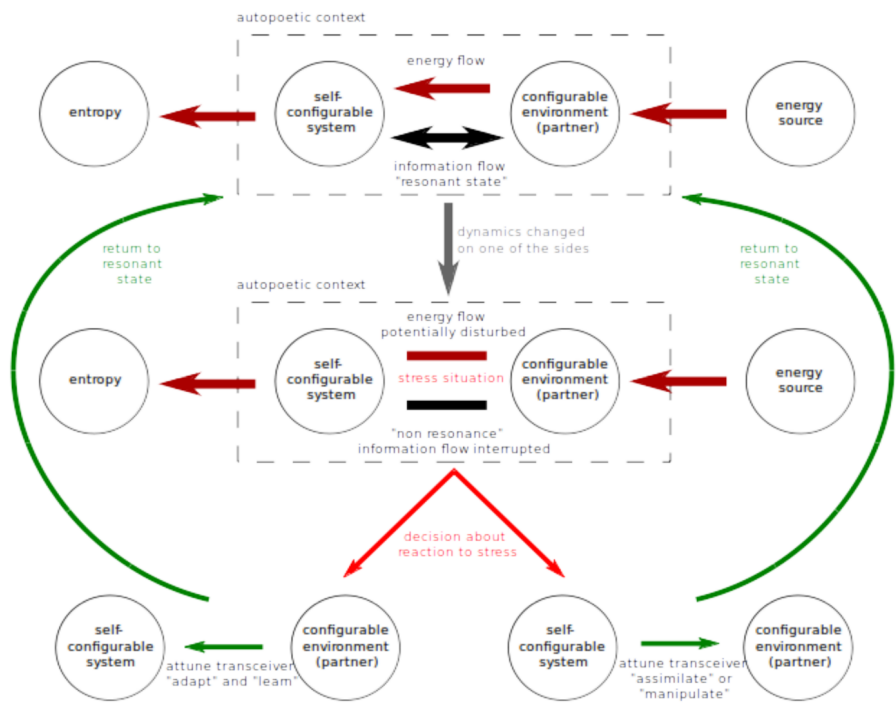

*Figure 62: One synchronization process can result either in a complete system update (full reconfiguration) or in a complete environment update (controlled manipulation, action).*

## 7.5 Summary of Chapter

In this chapter a round about was given how self-configuring systems relate or produce behavior. A static configuration of a dynamic plant can produce sequences of static plant configurations. Any plant can be self-configuring which results in cascades of self-configurable systems which all create sequences by the same principle. Plants with static policies (final plants) will produce static plant configurations (output configurations) from dynamic input configuration which is a necessary type of plant in order to terminate the configuration-behavior conversion cascade.

Systems with reconfiguration capabilities can self-propel themselves in the pursuit for optimizing values reported over the monitoring channel $q$. However, changes to the environment or changes to storage system holding the reference configurations can excite this process. At worst, the resulting behaviors can be difficult to associate with any particular configuration in the storage.

This chapter also discussed the potential benefits of modal reconfiguration styles. Three potential benefits were identified: Cybernetic decoupling of dynamics between plant and configurator, optimization of internal bandwidths and improving system's reliability. Especially, the last item was discussed in light of an informational safety concept.

Finally, a brief comment was made on how self-configuring systems could decide to allocate cost of reconfiguration to the outside of the system: The system can store and choose configurations which will require more change on the outside than on the inside. More technically speaking, the controller attempts to minimize the amount of internal reconfiguration activities, for example because of limited bandwidths. As a result, a self-configuring system will prefer modification of external configurations as a side-effect of optimizing its reconfiguration strategy α. Consequently, a self-configurable system has no dedicated functionality for external action and internal reconfiguration but only one integrated mechanism which expresses itself in various ways depending on the situation and content of Φ.

# 8 Conclusions for Designers

## 8.1 Spaces

In this paper I discussed various questions of design and function associated with self-configuring capabilities and to get convinced that it is not completely unsound to treat all configuration problems as "finding a point in a configuration space".

Those spaces can suffer from various kinds of limitations, such as stability of dimensional ordering, quantization of dimensions or configuration space fragmentation. However, I also pointed out how to overcome such problems.

Frankly, a plausible hypothetical ability to convert from such models into a unified framework is sufficient for a systemic theory of configuration because a technical conversion is not necessary. The system designer need not convert any models, only architectural insights generated in the abstract model back into the technical domain at hand.

I have explored various basic types of ways how configurations can be formulated and this paper tried to argue that it is possible to convert many, if not all, formalisms into a single conceptual framework based on configuration spaces comprised of configuration spots – an idea strongly employed in pattern recognition (pattern space) and cybernetics (where it is the state space). I have advertised this idea because not all technical disciplines are relying on this kind of approach.

## 8.2 Storage

The source or pool of configurations is the storage. In several places it was signaled that such a storage can be understood verbatim or as abstract, contextually distorted space of configurations from which the controller will draw the configurations.

The conceptual design of the configuration source used in this paper is indeed more akin to computer memory. Access consists of an addressing and retrieval phase where bandwidth limitations limit the effective number and the size of configurations in that memory.

Organization of memory can be optimized in terms of vector-like decomposition, amount of shared information, compression and addressing patterns.

### 8.3 Configurable Plants

The relationship between static configurations and behaviors of dynamic plants was discussed on the grounds of a worst-case plant which is not technically feasible but has easy to analyze structure. A recursive argument was made that configurable plants can be nested until final plants implement static policies. This allows to cluster sequences of configurations into meta-configurations. Indeed, the reverse process has been discussed in which an initially monolithic plant was decomposed into smaller ones. Original function and decomposed function fragments can exist in the same plant if the size of $\Psi$ configurations is extended.

Decomposing monolithic functions into smaller fragments raises the problem how to connect the program paths gracefully and to avoid "hanging". The solution seems to lie in generalization of program paths towards more generic transformation fields which can be later consolidated and reused. However, these features are not described as duties of the reconfiguration controller and will require model extensions.

### 8.4 Configuration Controllers

There seem to be several basic ways how new configurations can be found. Configurations can be randomly invented, assembled, linearly interpolated, obtained through optimization or deduced.

In this process several communication channels and their performance characteristics influence the reconfiguration speed (cf. figure 38). It was discussed by which techniques the total speed could be increased by looking at each channel ($r$, $n$, $m$, $q$) individually.

This paper has discussed the idea of *real time* and *virtual time* and the potential treatment of problems occurring in this context. One such problem could be that a system endangers its existence if it cannot promptly respond to environmental input.

Implementing slow spontaneous or swiftly iterative reconfiguration has effects on viability of certain reconfiguration strategies. In fact, most reconfiguring systems implement spontaneous configuration transitions either because they can be taken offline at scheduled times or because the interruption of service is not critical.

However, more and more products with embedded computers enter the market. Those systems have dynamic constraints which they must respect. Here the reconfiguration process can become an integral part of system's advertised behavior. In that context, already small differences in technical performance can express themselves to the system as landscapes of reconfigurational Nash-equilibria which can lead to significantly different system behaviors ("evasive behaviors"). This insight is used to motivate exploitation of all reconfiguration tricks to make transitions between configurations as fast as possible.

### 8.5 Ω-Units

This paper was motivated by the question if external configuration and self-configuration can be understood in a generic, systemic way, so that self-configuring networks, configurable computing platforms and robotic appliances can be understood in a single, generic framework – a theory of configuration.

For that reason I have chosen a basic model to describe (self-)configurable systems which is mainly relying on link speeds ("bandwidths") between the three basic parts *storage*, *configurator* and *plant*. This choice has high chances to be applicable across many different kinds of systems. From this model I derived several basic approaches how to improve exploitation of existing bandwidths. The solutions should look familiar to technological solutions in telecommunications, computing, AI or robotics. I have drawn attention to similarities with those domains wherever it deemed possible.

Additionally, elements and manipulation of self-configuring systems architectures was discussed in the hope that it is now easier to detect systems with self-configuring capabilities (cf. figure 39).

## 8.6  Further Work

Furthermore, this paper has identified at least two areas for further work:

1) The concept of hazard keys and informational safety / product safety

2) A configuration refactorization theory based on sketches of the Ω-transformations. Maybe it is possible to expand this concept towards an algebraic system.

There are several other issues which will be investigated in future work:

- What are the limits in converting explicit optimization parameters in controllers into implicit parameters expressed as storage space organization?

- Given current level of control theory for linear systems, how well can this be converted $\Omega_{min}$-units?

- Which general purpose optimization criteria exist for a configuration controller which are not part of the storage structure and need not to be provided externally?

- Obviously, learning and reuse are not included in the theory. How can this model be expanded in order to accommodate explanation of learning and production of universal operations?

## 8.7  Utility & Applicability

The theoretical considerations for self-configurable systems which were developed and explored in this paper have an overview character and hence should be valuable to system designers as a guiding instrument: Any designer dealing with a new system implementation requiring self-configuring capabilities in his system can systematically explore optimization potentials based on the three main communication channels $n$, $m$, $q$ and $r$ and eventually rely on standard propositions how to improve their performance.